\documentclass{article}
\pdfoutput=1
\usepackage{ijcai18}

\usepackage{times} 
\usepackage{soul} 
\usepackage[utf8]{inputenc} 
\usepackage[small]{caption} 

\usepackage{graphicx}  
\usepackage{subcaption} 
\usepackage{amsmath,amsthm,amssymb}
\usepackage{thm-restate}
\usepackage{booktabs} 
\usepackage{standalone} 

\usepackage{tikz}
\usetikzlibrary{arrows,positioning,decorations.pathreplacing}

\usepackage[bookmarks=true]{hyperref}
\hypersetup{
  pdfinfo={
    Author={William Vega-Brown and Nicholas Roy},
    Title={Admissible abstractions for near-optimal task and motion planning},
    CreationDate={D:20180130120000},
    Subject={Integrated task and motion planning},
    Keywords={task and motion planning;angelic;abstraction},
  }
}

\usepackage{algorithm} 
\usepackage[noend]{algpseudocode}

\newtheorem{theorem}{Theorem}
\newtheorem{definition}{Definition}
\newtheorem{corollary}{Corollary}
\newtheorem{proposition}{Proposition}

\makeatletter
\algrenewcommand\ALG@beginalgorithmic{\small}
\def\ALG@step%
   {%
   \refstepcounter{ALG@line}
   \stepcounter{ALG@rem}
   \ifthenelse{\equal{\arabic{ALG@rem}}{\ALG@numberfreq}}%
      {\setcounter{ALG@rem}{0}\alglinenumber{\arabic{ALG@line}}}%
      {}%
   }%
\newcounter{algorithmicH}
\let\oldalgorithmic\algorithmic
\renewcommand{\algorithmic}{%
	\stepcounter{algorithmicH}
	\oldalgorithmic}
	\renewcommand{\theHALG@line}{ALG@line.\thealgorithmicH.\arabic{ALG@line}}
\makeatother

\newcommand{\AbstractOperator}{\mathbf{a}}
\newcommand{\AbstractPlan}{\mathbf{p}}
\newcommand{\PrimitivePlan}{p}
\newcommand{\PrimitivePlans}{\mathcal{P}}
\newcommand{\PrimitiveOperator}{a}
\newcommand{\PrimitiveOperators}{\mathcal{A}_0}
\newcommand{\AbstractState}{\mathbf{s}}
\newcommand{\PrimitiveState}{x}
\newcommand{\Reals}{\mathbb{R}}
\newcommand{\Integers}{\mathbb{Z}}

\newcommand{\NonnegativeReals}{\mathbb{R}_{\ge0}}
\newcommand{\TopLevelOperator}{\textsc{Act}}

\title{Admissible Abstractions for Near-optimal Task and Motion Planning\thanks{%
    This document is an extended version of \protect\cite{vega-brown2016admissible} containing additional proofs and exposition.}
}

\author{
  William Vega-Brown \and Nicholas Roy \\
  Massachusetts Institute of Technology \\
  \{wrvb, nickroy\}@mit.edu
}

\begin{document}

\maketitle

\begin{abstract}
  We define an admissibility condition for abstractions expressed using angelic semantics and show that these conditions allow us to accelerate planning while preserving the ability to find the optimal motion plan.
  We then derive admissible abstractions for two motion planning domains with continuous state.
  We extract upper and lower bounds on the cost of concrete motion plans using local metric and topological properties of the problem domain.
  These bounds guide the search for a plan while maintaining performance guarantees.
  We show that abstraction can dramatically reduce the complexity of search relative to a direct motion planner.
  Using our abstractions, we find near-optimal motion plans in planning problems involving $10^{13}$ states without using a separate task planner.
\end{abstract}

\section{Introduction}

\begin{figure}
  \centering
    \begin{subfigure}[t]{0.5\columnwidth}
      \centering
      \includegraphics[width=\textwidth]{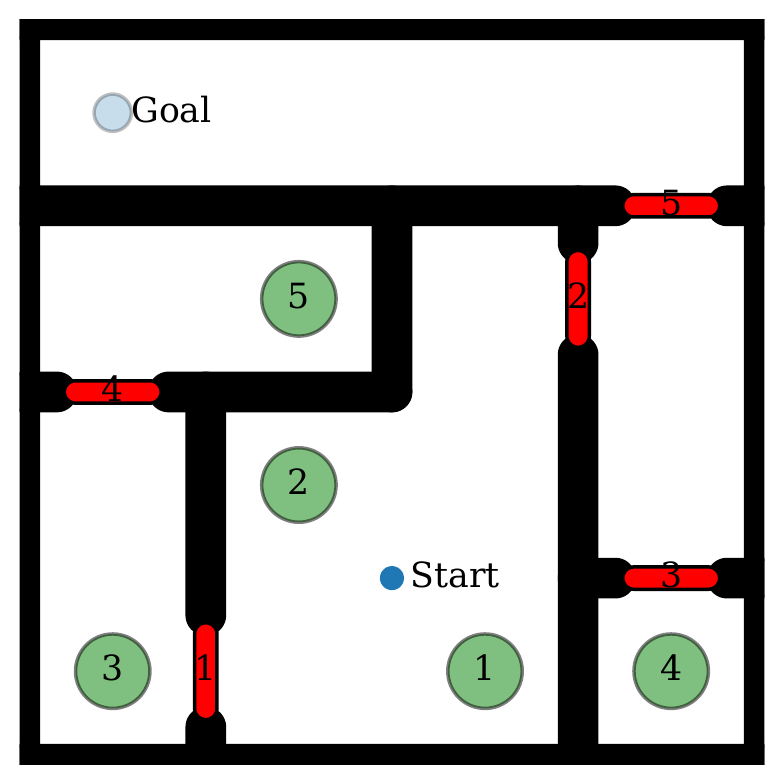}
      \caption{The door puzzle problem}
      \label{fig:example_problem}
    \end{subfigure}%
    \begin{subfigure}[t]{0.5\columnwidth}
      \centering
      \includegraphics[width=\textwidth]{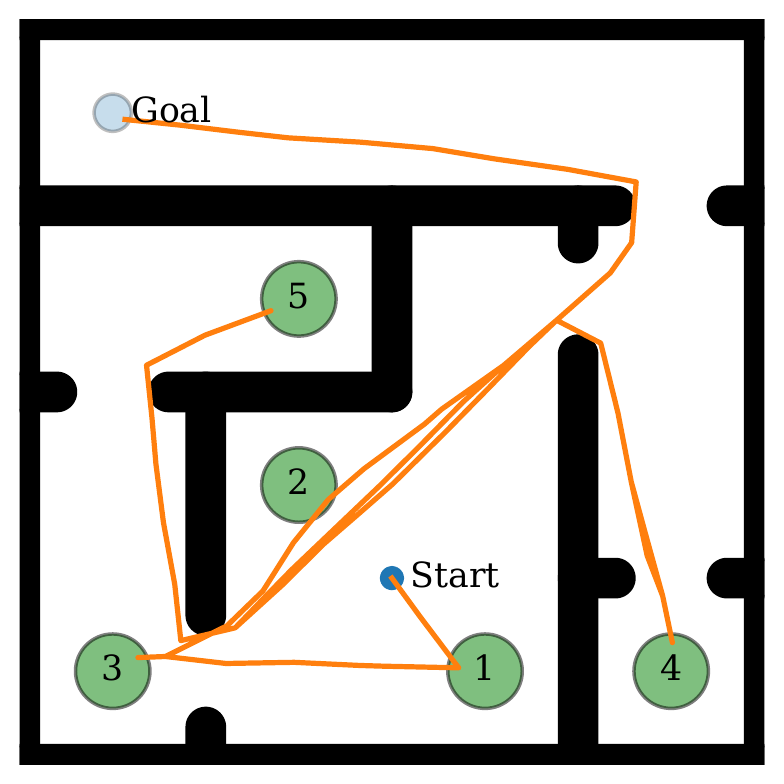}
      \caption{The optimal solution}
      \label{fig:example_solution}
    \end{subfigure}
  \caption{%
    The door-switch problem, an example task and motion planning domain.
    A two-dimensional robot must navigate from the start location to a goal location, but the way is obstructed by doors that can only be opened by toggling a corresponding switch.
    The optimal solution to this problem instance is to toggle the switches in the order $(1, 3, 2, 4, 5)$ and then go to the goal set.
    Because the size of the configuration space grows exponentially with the number of doors, planning is computationally challenging.
    Abstraction can render such planning problems tractable.
    \label{fig:example}
  }
\end{figure}

Consider a problem domain like the one shown in figure~\ref{fig:example}.
A holonomic two-dimensional agent is tasked with navigating to a specified goal region as quickly as possible.
The path is blocked by doors that can only opened by pressing the appropriate switch.
Planning the sequence of switches to toggle requires combinatorial search; deciding if a path exists to each switch requires motion planning.
As in many real-world planning domains, such as object manipulation or navigation among movable objects, the combinatorial search and motion planning problems are coupled and cannot be completely separated.

A standard approach to making such problems computationally tractable is to use abstraction to reason about the properties of groups of primitive plans simultaneously.
For example, we could choose a sequence of high-level operations using a task planner, ignoring the details of the underlying motion plan.
If we later determine that we cannot find a motion plan consistent with our high-level plan, we can use that information to modify our high-level plan.
For example, \cite{gravot2005asymov} describe an integrated approach that relies on a heuristic search for a high-level plan and uses motion planners as subroutines to deal with detailed geometry.
\cite{kaelbling2011hierarchical} use a hierarchy to guide high-level decision making, resolving low-level decisions arbitrarily and trusting in the reversibility of the system to ensure hierarchical completeness. 
Although these and other approaches (e.g.,~\cite{garrett2015ffrob,cambon2009hybrid,srivastava2013using}) vary in how they deal with the interaction between geometric planning and combinatorial search, they share a common weakness: they can only make guarantees about the plans they generate relative to the abstraction they are provided.
Even optimizing approaches (\cite{wolfe2010combined}) are generally limited to guarantees of hierarchical optimality.

Angelic semantics (\cite{marthi2008angelic}) provide a way to describe an abstraction that preserves optimality, but it is not clear what criteria an angelic abstraction must satisfy in order to make guarantees about the quality of synthesized plans.
In this paper, we describe conditions under which an abstraction will preserve the ability to find the optimal motion plan while accelerating planning.
We derive abstractions for two continuous planning domains, and using these abstractions we can dramatically reduce the complexity of search relative to a direct motion planner.
We find near-optimal motion plans in planning problems involving $10^{13}$ states without using a separate task planner.

\section{Problem Formulation}

We are interested in planning problems involving some underlying continuous configuration space $\mathcal{X}$, such as the position of a robot or the configuration of its joints.
Our task is to find a path through free space that starts in a specified state $s_0$ and ends in a goal set $S_{\mathrm{goal}}$.
This goal set may be specified implicitly, as the set of all states satisfying some constraint.

A path is a continuous map $\PrimitivePlan:[0, 1]\to \mathcal{X}$.
We define a concatenation operator $\circ$ for paths.
\begin{equation}
  (p_1\circ p_2)(s) =
  \begin{cases}
    p_1(2t) & \mathrm{if\,} t \le \frac{1}{2} \\
    p_2(2t-1) & \mathrm{if\,} \frac{1}{2} < t \le 1
  \end{cases}
\end{equation}
Let $\PrimitivePlans_{\mathcal{X}}(S, S')$ be the set of all paths starting in $S \subset \mathcal{X}$ and ending in $S' \subset \mathcal{X}$.
Let $c: \mathcal{X} \times T\mathcal{X} \to \mathbb{R}_{>0}$ be a cost function, where $T\mathcal{X}$ is the tangent space of $\mathcal{X}$.
We can define an associated cost functional $\mathcal{C}:P \to \mathbb{R}_{\ge 0}$.
\begin{equation}
  \mathcal{C}[\PrimitivePlan] = \int_0^1 c(\PrimitivePlan(t), \dot{\PrimitivePlan}(t))\,\mathrm{d}t
\end{equation}
Because $\mathcal{C}$ is additive, $\mathcal{C}[p_1 \circ p_2] = \mathcal{C}[p_1] + \mathcal{C}[p_2]$.
We define the \emph{optimal cost function} $c^*: 2^{\mathcal{X}} \times 2^{\mathcal{X}} \to \mathbb{R}_{\ge 0}$ as
\begin{equation}
  c^*(S, S') = \inf \{\mathcal{C}(\PrimitivePlan): \PrimitivePlan \in \PrimitivePlans_{\mathcal{X}}(S, S')\}.
\end{equation}

We define the $\epsilon$-approximate planning problem as the search for a path $\hat{\PrimitivePlan} \in \PrimitivePlans_{\mathcal{X}}(\{s_0\}, S_g)$ with cost less than $(1+\epsilon)$ the optimal cost for any $\epsilon \in \mathbb{R}_{\ge 0} \cup \{\infty\}$.
\begin{equation}
  \hat{\PrimitivePlan} \in \{ \PrimitivePlan \in \PrimitivePlans_{\mathcal{X}}(\{s_0\}, S_g) : \mathcal{C}(\hat{\PrimitivePlan}) \le (1 + \epsilon) c^*(s_0, S_g) \}
\end{equation}
The case where $\epsilon=\infty$, when we wish to find any feasible path to the goal set, is the problem of \emph{satisficing} planning.
The case where $\epsilon=0$ is optimal planning.

The set $\PrimitivePlans_{\mathcal{X}}(\mathcal{X}, \mathcal{X})$ of all possible paths from all possible start and goal locations is continuous and topologically complex.
To simplify planning, we assume we have available a finite set $\mathcal{A}_0$ of \emph{primitive operators}, low-level actions that can be executed in the real world.
The problem of constructing such a set of operators in continuous motion planning domains is well studied; in this document, we will assume the set of operators are given by the edges in a probabilistic roadmap (PRM*).
That is, we randomly sample a finite set of configurations $\mathcal{V}_n \subset \mathcal{X}$, and for each such configuration $v$, we define an operator $p_v$.
The operator $p_v$ ensures that the robot will end at the state $v$ if executed from any state in the open ball of radius $r_n$ around $v$, where $r_n \propto (\log n / n)^{1/d}$ is a radius that increases slowly with the size of the discretization.
Any feasible plan can be well-approximated by a sequence of these randomly sampled operators as the number of sampled configurations tends to infinity.
For example, we can show that if $\mathcal{A}_{0,n}^*$ is the set of all paths through a PRM* with $n$ sampled configurations, then
\begin{multline}
  \label{eq:asymptotically_optimal}
  \lim_{n\to\infty} \inf \{\mathcal{C}[\PrimitivePlan]: \PrimitivePlan \in \mathcal{A}_{0,n}^* \cap \PrimitivePlans_{\mathcal{X}}(\{s_0\}, S_g)\} = \\
  \inf \{\mathcal{C}[\PrimitivePlan]: \PrimitivePlan \in \PrimitivePlans_{\mathcal{X}}(\{s_0\}, S_g)\}.
\end{multline}
This was proven by \cite{karaman2011sampling} for the case where the system is subject to analytic differential constraints, and by \cite{vega-brown2016asymptotically} when the system has piecewise-analytic differential constraints (as in object manipulation problems).

Because the set of primitive operators can grow quite large, especially in problems with high-dimensional configuration spaces, a direct search for primitive plans is computationally intractable.
Instead, we will use angelic semantics to encode bounds on the cost of large groups of plans.
We can use these bounds to plan efficiently while preserving optimality.

\section{Angelic Semantics}

\begin{figure*}
  \centering
    \begin{subfigure}[t]{0.5\textwidth}
      \centering
      \includegraphics[width=\textwidth]{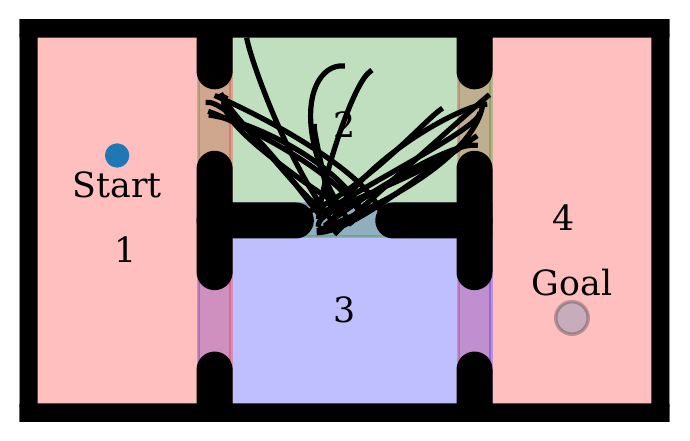}
      \caption{Plans in operator $\AbstractOperator_{23}$}
      \label{fig:abstraction:operator}
    \end{subfigure}%
    \begin{subfigure}[t]{0.5\textwidth}
      \centering
      \includegraphics[width=\textwidth]{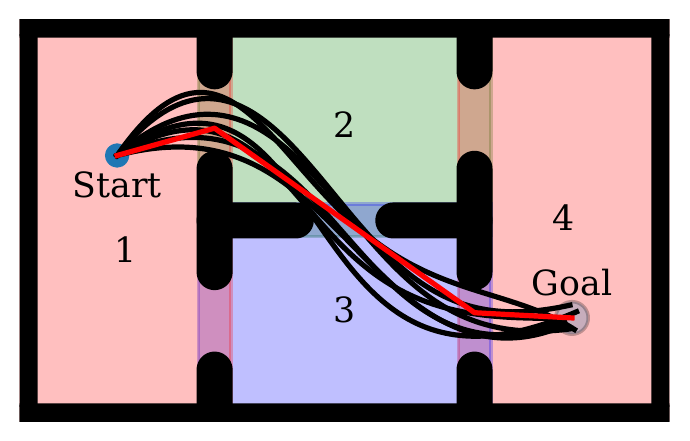}
      \caption{Plans in $\AbstractOperator_{12} \circ \AbstractOperator_{23} \circ \AbstractOperator_{34} \circ \AbstractOperator_{4g}$}
      \label{fig:abstraction:plan}
    \end{subfigure}
    \begin{subfigure}[t]{0.5\textwidth}
      \centering
      \includegraphics[width=\textwidth]{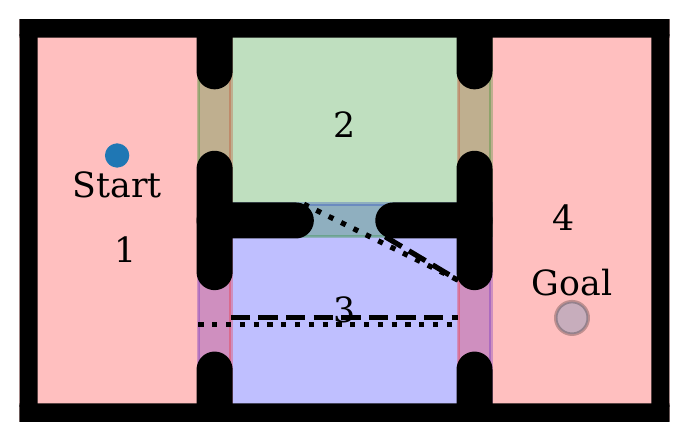}
      \caption{Bounds for $V[\AbstractOperator_{34}]$}
      \label{fig:abstraction:valuation}
    \end{subfigure}%
    \begin{subfigure}[t]{0.5\textwidth}
      \centering
      \includegraphics[width=\textwidth]{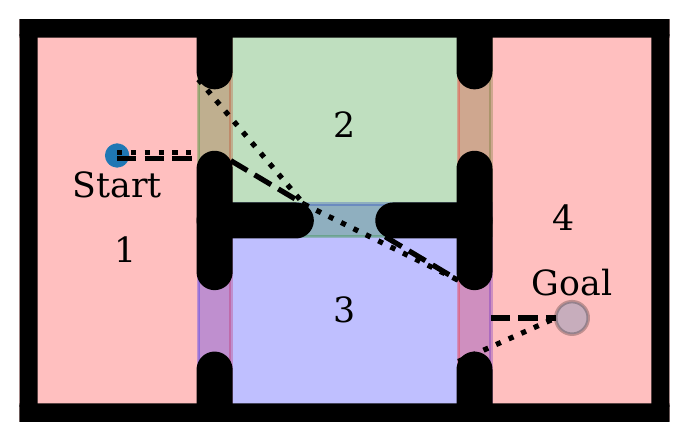}
      \caption{Bounds for $V[\AbstractOperator_{12} \circ \AbstractOperator_{23} \circ \AbstractOperator_{34} \circ \AbstractOperator_{4g}]$}
      \label{fig:abstraction:propagation}
    \end{subfigure}
    \begin{subfigure}[t]{0.5\textwidth}
      \centering
      \includegraphics[width=\textwidth]{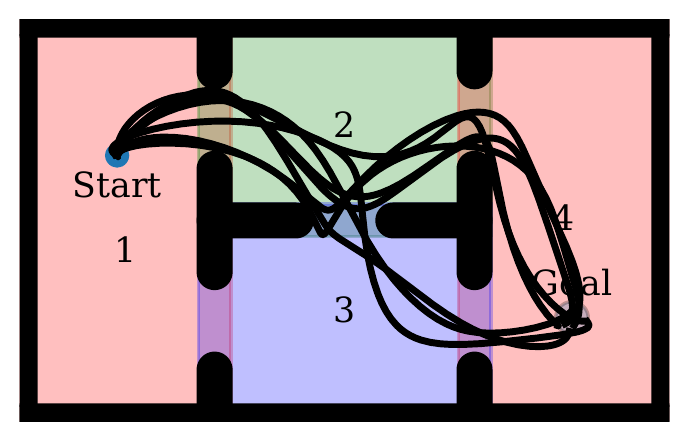}
      \caption{Plans in $\AbstractOperator_{12} \circ \TopLevelOperator$}
      \label{fig:abstraction:refinement_a}
    \end{subfigure}%
    \begin{subfigure}[t]{0.5\textwidth}
      \centering
      \includegraphics[width=\textwidth]{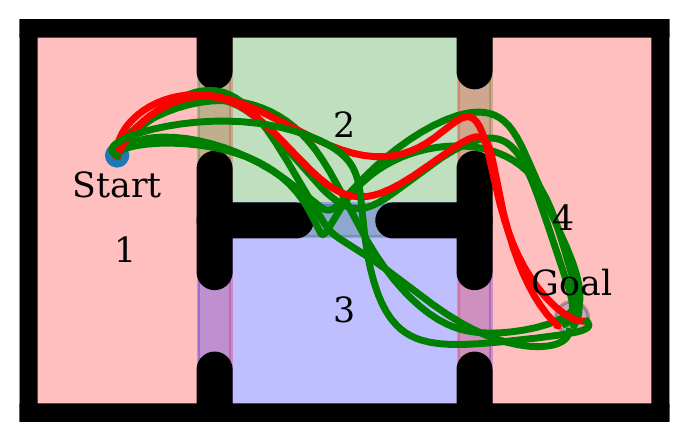}
      \caption{Plans in $\AbstractOperator_{12} \circ \AbstractOperator_{23} \circ \TopLevelOperator$}
      \label{fig:abstraction:refinement_b}
    \end{subfigure}
  \caption{%
    A schematic description of angelic semantics.
    Abstract operators (a) are sets of primitive plans, and can be defined implicitly in terms of constraints.
    For example, the operator $\AbstractOperator_{23}$ contains all plans that end in region $3$ and do not leave region $2$.
    We can sequence abstract operators into abstract plans (b).
    The red lines link the centroids of successive regions, while the black lines are randomly sampled primitive plans representative of the abstract plan that move from regions $1$ to $2$ to $3$ to $4$ to the goal.
    We can use domain-specific information to compute bounds on the cost of any plan in an operator stating from a specific set of states (c).
    Here, lower bounds are drawn using dashed lines, while upper bounds are drawn in dotted lines.
    Note the dependence on the initial state: the cost of a plan starting in $1 \cap 3$ is strictly higher than the cost of a plan stating in $2 \cap 3$.
    We can sequence these bounds (d) to compute bounds on the cost of an abstract plan.
    Finally, a refinement of an abstract plan $\AbstractPlan$ (e) is a less abstract plan (f) $\AbstractPlan' \subset \AbstractPlan$.
    Primitive plans in $\AbstractPlan'$ are shown with heavy lines, while plans in $\AbstractPlan \setminus \AbstractPlan'$ are shown with finer lines.
    \label{fig:abstraction}}
\end{figure*}

An \emph{abstract operator} $\AbstractOperator$ represents a set $\AbstractOperator \subset \PrimitivePlans_{\mathcal{X}}$ of primitive plans.
Because the space of plans is infinite, we define operators implicitly, using constraints on the underlying primitive plans.
For example, in a navigation problem, we might define an operator as any primitive plan that remains inside a given set of configuration space and ends in a different set of configuration space.
This is depicted graphically in figure~\ref{fig:abstraction:operator}: the operator $\AbstractOperator_{23}$ contains every path that is contained in region $2$ and ends in region $3$.

The concatenation of two operators $\AbstractOperator_i \circ \AbstractOperator_j$ is an abstract plan containing all possible concatenations of primitive plans in the operators.
\begin{equation}
  \AbstractOperator_i \circ \AbstractOperator_j = \{
    p_i \circ p_j : p_i \in \AbstractOperator_i, p_j \in \AbstractOperator_j, p_i(1) = p_j(0)
  \}
\end{equation}
The condition $p_i(1) = p_j(0)$ is necessary to enforce that only feasible plans are contained in $\AbstractOperator_i \circ \AbstractOperator_j$.
In a problem with nontrivial dynamic constraints, the condition would need to be more complex.
In figure~\ref{fig:abstraction:plan}, we show samples from the plan $\AbstractOperator_{12} \circ \AbstractOperator_{23} \circ \AbstractOperator_{34} \circ \AbstractOperator_{4g}$, which contains paths that move from region 1 to 2 to 3 to 4 to the goal.
The concatenation operation allows us to express complicated sets of plans in a compact way.

Because our operators are defined implicitly, it can be difficult to find the best plan in the abstract plan, or even to decide if there exists a plan consistent with the constraints of an abstract plan.
Note that it is easy to write down abstract plans that are empty; in the toy navigation example in figure~\ref{fig:abstraction}, the plan $\AbstractOperator_{12} \circ \AbstractOperator_{34}$ contains no primitive plans, as the intersection of regions $1$, $2$, and $3$ is empty.
For planning with abstract operators to be feasible, we need a way to reason about the primitive plans contained by an abstract plan \emph{without} first enumerating those primitive plans.

Specifically, we will develop a way to compare abstract plans, and we will show this comparison is sufficient for planning.
We do this using the \emph{valuation} of an operator or plan.
A \emph{valuation} $V[\AbstractOperator]$ for an operator or plan $\AbstractOperator$ is the unique map $V[\AbstractOperator]: \mathcal{X} \times \mathcal{X} \to \mathbb{R}_{\ge 0}$ that takes a pair of states and gives the lowest cost path between the pair.
\begin{equation}
  V[\AbstractOperator](s_1, s_2) = \inf \{\mathcal{C}(\sigma):
  \sigma\in \AbstractOperator, \sigma(0)=s_1, \sigma(1)=s_2 \} \label{eq:valuation}
\end{equation}
Note that if there are no paths in $\AbstractOperator$ linking $s_1$ and $s_2$, then $V[\AbstractOperator](s_1, s_2) = \inf \varnothing = \infty$.

Valuations allow us to compare abstract plans without reference to the primitive plans they contain.
Given two abstract plans $\AbstractPlan$ and $\AbstractPlan'$, if we can prove that for any pair of states $\PrimitiveState, \PrimitiveState'$, either $V[\AbstractPlan](\PrimitiveState, \PrimitiveState') < V[\AbstractPlan'](\PrimitiveState, \PrimitiveState')$ or $V[\AbstractPlan'](\PrimitiveState, \PrimitiveState') = \infty$, then either there is a solution to our planning problem in $\AbstractPlan$, or there is no solution in $\AbstractPlan$ or $\AbstractPlan'$.
Either way, we do not need to consider any plan in $\AbstractPlan'$; we can prune $\AbstractPlan'$ from our search space.
Under such a condition, we say that $\AbstractPlan$ \emph{dominates} $\AbstractPlan'$ and we write $V[\AbstractPlan] \prec V[\AbstractPlan']$.
Similarly, if either  $V[\AbstractPlan](\PrimitiveState, \PrimitiveState') \le V[\AbstractPlan'](\PrimitiveState, \PrimitiveState')$ or $V[\AbstractPlan'](\PrimitiveState, \PrimitiveState') = \infty$, then we say that $\AbstractPlan$ \emph{weakly dominates} $\AbstractPlan'$ and we write $V[\AbstractPlan] \preceq V[\AbstractPlan']$.

Unfortunately, determining the valuation of an operator is itself an optimization problem, and one that is not necessarily any easier than the planning problem we are trying to solve.
In many cases, however, we can derive a computational advantage from reasoning about \emph{bounds} on the valuation of an abstract operator.
By representing these bounds \emph{symbolically}, we will be able to reason without reference to the underlying states or plans.

We first define bounds on the valuation of an operator over a set of states.
\begin{align}
  V_L[\AbstractOperator](\AbstractState, \AbstractState') = \inf \{ \inf \{ V[\AbstractPlan](s, s'): s' \in \AbstractState' \}: s \in \AbstractState \} \\
  V_U[\AbstractOperator](\AbstractState, \AbstractState') = \sup \{ \inf \{ V[\AbstractPlan](s, s'): s' \in \AbstractState' \}: s \in \AbstractState \}
\end{align}
A symbolic valuation bound $\hat{V}[\AbstractOperator]$ is a set of tuples $\{(\AbstractState, \AbstractState', l, u)\}$, where $\AbstractState, \AbstractState'$ are symbolic states and $l < u \in \NonnegativeReals \cup \{\infty\}$.
A bound $\hat{V}[\AbstractOperator]$ is admissible if
\begin{align}
  \exists (\AbstractState, \AbstractState', l, u) &\in \hat{V}[\AbstractOperator]:&
  l &\le V_L[\AbstractOperator](\AbstractState, \AbstractState') \\
  \forall (\AbstractState, \AbstractState', l, u) &\in \hat{V}[\AbstractOperator]:&
  u &\ge V_U[\AbstractOperator](\AbstractState, \AbstractState').
\end{align}
In words, a bound $(\AbstractState, \AbstractState', l, u)$ is admissible if for any state $\PrimitiveState$ in $\AbstractState$ there exists a plan $\PrimitivePlan$ ending in some state $\PrimitiveState'$ in $\AbstractState'$ with cost $c = \mathcal{C}[p]$ bounded above by $u$ and below by $l$.
We can also interpret a symbolic valuation bound $\hat{V}$ as a bound over sets of states.
\begin{align}
  \hat{V}_L[\AbstractOperator](\AbstractState, \AbstractState') =&
  \inf \{l: (\AbstractState_0, \AbstractState_1, l, u) \in \hat{V}[\AbstractOperator], \AbstractState \cap \AbstractState_0 \ne \varnothing, \nonumber\\
    &\qquad\qquad \AbstractState' \cap \AbstractState_1 \ne \varnothing\} \\
    \hat{V}_U[\AbstractOperator](\AbstractState, \AbstractState') =&
    \inf \{u: (\AbstractState_0, \AbstractState_1, l, u) \in \hat{V}[\AbstractOperator], \AbstractState \subseteq \AbstractState_0, \AbstractState' \subseteq \AbstractState_1\}.
\end{align}
Note that if $\hat{V}[\AbstractOperator]$ is admissible, then $\hat{V}_L[\AbstractOperator](\AbstractState, \AbstractState') \le V_L[\AbstractOperator](\AbstractState, \AbstractState')$ and $V_U[\AbstractOperator](\AbstractState, \AbstractState') \le \hat{V}_U[\AbstractOperator](\AbstractState, \AbstractState')$ for all abstract state pairs $\AbstractState, \AbstractState'$ (see appendix~\ref{sec:proofs}, proposition~\ref{thm:bounds}).

This observation has important consequences in a few interesting limiting cases.
A bound $\hat{V}[\AbstractOperator]$ contains at least one element $(\AbstractState, \AbstractState', l, u)$ where $u$ is finite only if then there must be some plan in the operator $\AbstractOperator$.
A bound $\hat{V}[\AbstractOperator]$ does not contain an element $(\AbstractState, \AbstractState', l, u)$ where $l$ is finite only if $\AbstractOperator$ is empty.
Similarly, $\hat{V}_U[\AbstractOperator](\AbstractState, \AbstractState') < \infty$ implies $\AbstractOperator$ contains feasible plans connecting each state in $\AbstractState$ to some state in $\AbstractState'$, while $\hat{V}_L[\AbstractOperator](\AbstractState, \AbstractState') = \infty$ implies $\AbstractOperator$ contains no plan connecting a state in $\AbstractState$ to a state in $\AbstractState'$.
In addition, if $\hat{V}[\AbstractPlan]$ and $\hat{V}[\AbstractPlan']$ are admissible, then $\hat{V}[\AbstractPlan \cup \AbstractPlan'] = \hat{V}[\AbstractPlan] \cup \hat{V}[\AbstractPlan']$ is admissible (see appendix~\ref{sec:proofs}, proposition~\ref{thm:join} for a proof).

It is also important to recognize the state-dependence of valuation bounds.
Consider the operator $\AbstractOperator_{34}$ in figure~\ref{fig:abstraction:valuation}; the operator is defined as containing any plan contained in region $3$ that ends in region $4$.
Because regions $3$ and $4$ intersect, the global lower bound on the cost of a plan in this operator is zero.
However, we can compute nontrivial bounds for specific states, or for specific sets of states.
For example, paths achieving lower and upper bounds are drawn from the abstract states $R_2 \cap R_3$ and $R_1 \cap R_3$ to the termination set of the operator.

As we will see in sections~\ref{sec:abstraction:navigation} and \ref{sec:abstraction:door_puzzle}, for many domains we will not need to write down a valuation explicitly.
Instead, we can use domain information to make metric computations and generate the necessary elements of a valuation procedurally.
Moreover, by working with symbolic bounds we can efficiently compute bounds on the cost of plans consisting of sequences of abstract operators, without reference to a dense discretization of the underlying space of plans.
For example, if we have bounds on a plan $\hat{V}[\AbstractOperator]$ and an operator $\hat{V}[\AbstractOperator']$, we can compute a bound $\hat{V}[\AbstractOperator \circ \AbstractOperator']$.
\begin{align}
  \hat{V}[\AbstractOperator \circ \AbstractOperator'] =
  &\{(\AbstractState, \AbstractState''', l+l', u+u'):
    (\AbstractState, \AbstractState', l, u) \in \hat{V}[\AbstractOperator],
    \nonumber \\&\quad
    (\AbstractState'', \AbstractState''', l', u') \in \hat{V}[\AbstractOperator'],
    \AbstractState' \subseteq \AbstractState''
  \} \,\cup \nonumber \\
  & \{(\AbstractState, \AbstractState''', l+l', u):
    (\AbstractState, \AbstractState', l, u') \in \hat{V}[\AbstractOperator],
    \nonumber \\&\quad
    (\AbstractState'', \AbstractState''', l', \infty) \in \hat{V}[\AbstractOperator'],
    \AbstractState' \cap \AbstractState'' \ne \varnothing
  \}
\end{align}
If $\hat{V}[\AbstractOperator]$ and $\hat{V}[\AbstractOperator']$ are admissible, then $\hat{V}[\AbstractOperator \circ \AbstractOperator']$ is admissible as well (see appendix~\ref{sec:proofs}, proposition~\ref{thm:propagation} for a proof).
We call this process \emph{propagation}.
This process is depicted graphically in figure~\ref{fig:abstraction:propagation}.

\section{Admissible Abstractions}
\label{sec:refinement}

We will use angelic semantics to specify abstractions that enable efficient planning.
Suppose that $\AbstractPlan, \AbstractPlan'$ are abstract plans, with $\AbstractPlan \subset \AbstractPlan'$.
Then $\AbstractPlan' \preceq \AbstractPlan$, since any plan in $\AbstractPlan$ is also in $\AbstractPlan'$---but because $\AbstractPlan$ is a smaller set than $\AbstractPlan'$, our bounds may tighter.
If $\hat{V}_U[\AbstractPlan'] \prec \hat{V}_L[\AbstractPlan]$, then we can also conclude that $\AbstractPlan' \setminus \AbstractPlan \prec \AbstractPlan'$.
We can incrementally construct an increasingly accurate estimate of $V[\AbstractPlan]$ by iteratively considering smaller and smaller subsets of an operator $\AbstractPlan$ and pruning those subsets that cannot contain an optimal plan.
This is depicted graphically in figures~\ref{fig:abstraction:refinement_a} and \ref{fig:abstraction:refinement_b}.

We can make precise the construction of these increasingly fine subsets by introducing a \emph{refinement relation} $\mathcal{R} \subset \mathcal{A}^* \times \mathcal{A}^*$, where $*$ denotes the Kleene closure.
The elements of $\mathcal{R}$ are ordered pairs $(\AbstractPlan, \AbstractPlan')$ such that $\AbstractPlan' \subset \AbstractPlan$.
We can construct a relation $\mathcal{R}$ by defining a procedure to generate plans $\AbstractPlan'$ given a plan $\AbstractPlan$.
First, define an operation $\textsc{Head}:\mathcal{A}^*\to \mathcal{A}$, which takes a plan $\AbstractPlan$ and selects a single operator $\AbstractOperator$ from it to replace with a more constrained refinement.
We then define operations $\textsc{Base}:\mathcal{A}^* \to \mathcal{A}^*$ and $\textsc{Ext}: \mathcal{A}^* \to \mathcal{A}^*$ that return the part of $\AbstractPlan$ before and after $\textsc{Head}(\AbstractPlan)$, respectively.
Together, the three operators split a plan $\AbstractPlan$ into three segments so that $\AbstractPlan = \textsc{Base}(\AbstractPlan) \circ \textsc{Head}(\AbstractPlan) \circ \textsc{Ext}(\AbstractPlan)$.
Finally, we define a domain-specific relation $\bar{\mathcal{R}} \subset \mathcal{A} \times \mathcal{A}^*$; this can be thought of as a function mapping an abstract operators to a set of abstract plans.
Then $(\AbstractPlan, \AbstractPlan') \in \mathcal{R}$ if and only if $\AbstractPlan' = \textsc{Base}(\AbstractPlan) \circ \AbstractPlan'' \circ \textsc{Ext}(\AbstractPlan)$ and $(\textsc{Head}(\AbstractPlan), \AbstractPlan'') \in \bar{\mathcal{R}}$.
If $(\AbstractOperator, \AbstractPlan) \in \bar{\mathcal{R}}$, we call $\AbstractPlan$ a refinement of $\AbstractOperator$; similarly, if $(\AbstractPlan, \AbstractPlan') \in \mathcal{R}$, we call $\AbstractPlan'$ a refinement of $\AbstractPlan$.

We can combine these elements into an \emph{abstraction} over a problem domain $(\mathcal{X}, c, s_0, S_g)$.
Formally, an abstraction is a tuple $(\mathcal{S}, \mathcal{A}, \bar{\mathcal{R}}, \hat{V})$, where
\begin{itemize}
  \item $\mathcal{S}$ is a collection of propositional symbols,
  \item $\mathcal{A}$ is a collection of operators, including a distinguished top-level operator $\mathrm{Act}$,
  \item $\bar{\mathcal{R}} \subset \mathcal{A} \times \mathcal{A}^*$ is a refinement relation, and
  \item $\hat{V}$ is a symbolic valuation bound.
\end{itemize}
The valuation bound encodes both the cost and the dynamics of our problem domain.
The refinement relation structures the space of abstract plans.

Angelic planning algorithms accept an abstraction as an argument in much the same way that the A* search algorithm \cite{hart1968formal} accepts a heuristic.
This raises an important question: under what circumstances will an abstraction $(\mathcal{S}, \mathcal{A}, \bar{\mathcal{R}}, \hat{V})$ allow us to find the optimal primitive plan for a domain $(\mathcal{X}, c, s_0, S_g)$, and to prove we have done so?
We will generalize the idea of an admissible heuristic to define an \emph{admissible} abstraction.
As we will show in section~\ref{sec:algorithms}, two properties suffice.
\begin{definition}
  \label{thm:admissibility}
  An abstraction $(\mathcal{S}, \mathcal{A}, \bar{\mathcal{R}}, \hat{V})$, defined over a planning domain $(\mathcal{X}, c)$, is admissible if 
  \begin{enumerate}
    \item For each abstract operator $\AbstractOperator \in \mathcal{A}$, for each primitive plan $\PrimitivePlan$ in $\AbstractOperator$, there is a refinement $\AbstractPlan$ of $\AbstractOperator$ such that $p \in \AbstractPlan$, i.e.,
      \begin{equation}
        \forall \AbstractOperator \in \mathcal{A}, \forall p \in \AbstractOperator, \exists (\AbstractOperator,\AbstractPlan) \in \bar{\mathcal{R}}: p \in \AbstractPlan.
      \end{equation}
    \item $\hat{V}$ is admissible, i.e., $\hat{V}_L[\AbstractPlan] \preceq V[\AbstractPlan] \preceq \hat{V}_U[\AbstractPlan]$ for each abstract operator $\AbstractPlan \in \mathcal{A}$.
  \end{enumerate}
\end{definition}

The first property ensures that we do not ``lose track'' of any primitive plans while refining a plan.
Plans are only removed from consideration when they are deliberately pruned.
The second property ensures that if abstract plans $\AbstractPlan, \AbstractPlan' \in P$, where $P$ is a collection of abstract plans, and $\hat{V}_U[\AbstractPlan] \prec \hat{V}_L[\AbstractPlan']$, then no optimal plan is in $\AbstractPlan'$ and thus the best plan in $p$ is also in the set $P' = P \setminus \{\AbstractPlan'\}$.
Taken together, these properties ensure that if $P'$ is the result of refining and pruning a collection of plans $P$, then for every plan in $P$ there is a plan that is no worse in $P'$.
If we start with the set $P_0 = \{\mathrm{Act}\}$, no sequence of refinement and pruning operations will discard an optimal solution.
This ensures completeness.
To construct planning algorithms, we simply need to choose an order in which to refine and prune, and keep track of bounds to know when we can terminate the search.

\subsection{The Flat Abstraction for Graph Search}

We illustrate the construction of an admissible abstraction with graph search.
Let $\mathcal{G} = (\mathcal{V}, \mathcal{E})$ be a graph, where each edge $e \in \mathcal{E}$ has an associated cost $c_e$.
Suppose our objective is to find the shortest path to a goal node $v_g \in \mathcal{V}$ and we have an admissible heuristic $h: \mathcal{V} \to \NonnegativeReals$.
Then the abstraction $\mathcal{A}_\mathcal{G} = (\mathcal{V}, \mathcal{E} \cup \{\TopLevelOperator\}, \hat{V}, \bar{\mathcal{R}})$ is admissible, where
\begin{itemize}
  \item $\hat{V}[\AbstractOperator_e] = \{(\{e_0\}, \{e_1\}, c_e, c_e)\}$, 
  \item $\hat{V}[\TopLevelOperator] = \{ (\{v\}, \{v_g\}, h(v, v_g'), \infty): v \in \mathcal{V})\}$
  \item $\bar{\mathcal{R}}$ is the union of $\{(\TopLevelOperator, e \circ \TopLevelOperator) \forall e \in \mathcal{E} \}$ and $\{(\TopLevelOperator, e) \forall e \in \mathcal{E}: e_1 = v_g \}.$
\end{itemize}
Admissibility of $\hat{V}$ follows immediately from the admissibility of $h$, and the admissibility of $\bar{\mathcal{R}}$ is easily proven.
By definition, any primitive plan $\PrimitivePlan$ is contained in $\TopLevelOperator$.
Every primitive plan in the abstract plan $\PrimitivePlan \circ \TopLevelOperator$ is of the form $\PrimitivePlan \circ \PrimitivePlan'$.
Suppose the first primitive operator in $\PrimitivePlan'$ is $e$.
For each such $\PrimitivePlan$ and $\PrimitivePlan'$, $(\TopLevelOperator, e \circ \TopLevelOperator) \in \bar{\mathcal{R}}$.
Therefore $\bar{\mathcal{R}}$ is admissible, and so $\mathcal{A}_\mathcal{G}$ is admissible.

This demonstrates that the machinery of angelic abstractions is at least as general as heuristics in graph search: every graph search problem can be reformulated as an abstract search, using the edges to define a refinement operation and an admissible heuristic to define lower bounds.
Often, however, we can use domain-specific information to devise even more informative abstractions.
In the remainder of this section, we will provide concrete examples of admissible abstractions for a pair of simple \emph{continuous} planning problems.

\subsection{An Abstraction for Navigation}
\label{sec:abstraction:navigation}

A common problem in robotics is navigating to some specified goal location in a structured environment.
Simple heuristics like the Euclidean distance to the goal work well in environments that are cluttered but largely unstructured, where the distance is a good proxy for the true cost.
In highly structured environments, however, the Euclidean distance can be quite a bad proxy for cost.
Consider the example in figure~\ref{fig:navigation}, in which the robot starts just on the other side of a wall from the goal.
Using A* with a Euclidean heuristic requires searching almost the entire space.

We can plan more efficiently by taking advantage of structure in the environment.
Suppose we have a decomposition of the environment into a finite set of overlapping regions, as in figure~\ref{fig:abstraction}, and we know which regions overlap.
These regions can be derived from a semantic understanding of the environment, such as rooms and doorways, or they can be automatically extracted using (for example) the constrained Delaunay triangulation.
Then any plan can be described by the sequence of regions it moves through.
We can use this region decomposition to define an abstraction.

Let $\mathcal{S} = \{R_i\}$, where $\cup_i R_i = \mathcal{X}$, and let $\mathcal{A} = \mathcal{A}_0 \cup \{\AbstractOperator_{ij}: R_i \cap R_j \ne \varnothing \} \cup \{\TopLevelOperator\}$, where $\PrimitivePlan \in \AbstractOperator_{ij}$ if $\PrimitivePlan(s)\in R_i \forall s\in[0, 1) \wedge \PrimitivePlan(1) \in R_j$.
The refinement can be defined as follows.
\begin{equation}
  \begin{aligned}
    \bar{\mathcal{R}} =  \bigcup_{ij}
    & \{(\TopLevelOperator, \AbstractOperator_{ij} \circ \TopLevelOperator), (\TopLevelOperator, \AbstractOperator_{ij}) \} \,\cup \\
    & \{(\AbstractOperator_{ij}, \PrimitiveOperator \circ \AbstractOperator_{ij}): \PrimitiveOperator(t) \in R_i \forall t\} \,\cup \\
    & \{(\AbstractOperator_{ij}, \PrimitiveOperator ): \PrimitiveOperator(t) \in R_i \forall t, \PrimitiveOperator(1) \in \mathrm{cl}(R_j) \}
  \end{aligned}
\end{equation}
We can use spatial indices like k-D trees and R-trees to quickly find the operators that are valid from a particular state.
It is straightforward to show this refinement relation is admissible (see appendix~\ref{sec:proofs}, proposition~\ref{thm:regions_admissible}).

If the cost function is path length, then we can compute bounds using geometric operations.
Executing the action $\AbstractOperator_{ij}$ from a state in $R_k \cap R_i$ would incur a cost at least as great as the set distance $\inf \{\Vert s - s' \Vert: s \in R_i \cap R_k, s' \in R_i \cap R_j\}$.
If the intersections between sets are small and well-separated, this lower bound will be an accurate estimate.
This has the effect of heuristically guiding the search towards the next region, allowing us to perform a search in the (small) space of abstract plans rather than the (large) space of primitive plans.
The Euclidean heuristic can deal with things like clutter and unstructured obstacles, while the abstraction can take advantage of structure in the environment.

Note that we have made no reference to the shape of the regions, nor even to their connectedness.
If regions can be disconnected, for instance by an obstacle, abstract operators can have no upper bound, which can lead the search to be inefficient.
On the other hand, if we require the regions to be convex, then executing the action $\AbstractOperator_{ij}$ from a state in $R_k \cap R_i$ would incur a cost no greater than the Hausdorff distance $d_H(R_i \cap R_k, R_i \cap R_j)$, where
\begin{equation}
  d_H(X, Y) = \max(\sup_{x \in X} \inf_{y \in Y}\Vert x - y \Vert, \sup_{y \in Y} \inf_{x \in X}\Vert x - y \Vert).
\end{equation}

\begin{figure}
  \centering
  \begin{subfigure}[t]{0.5\columnwidth}
    \centering
    \includegraphics[width=\textwidth]{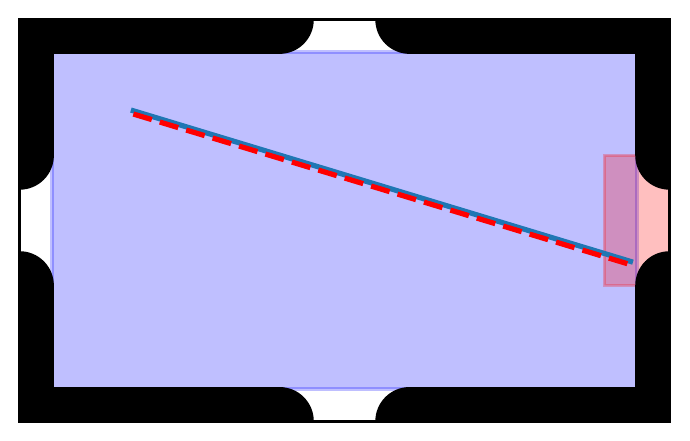}
    \caption{}
    \label{fig:convex:convex}
  \end{subfigure}%
  \begin{subfigure}[t]{0.5\columnwidth}
    \centering
    \includegraphics[width=\textwidth]{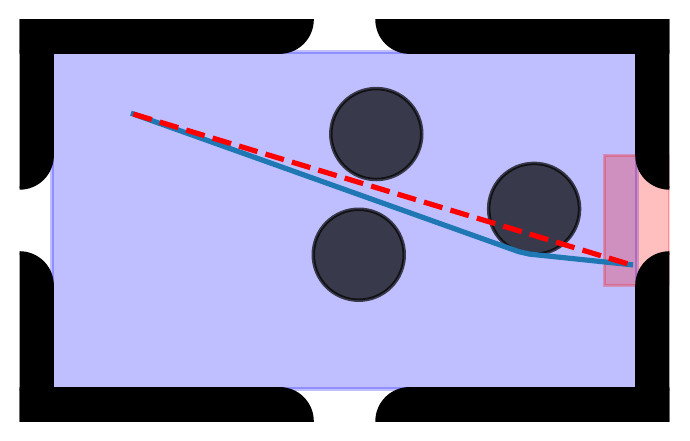}
    \caption{}
    \label{fig:convex:epsilon_convex}
  \end{subfigure}
  \begin{subfigure}[t]{0.5\columnwidth}
    \centering
    \includegraphics[width=\textwidth]{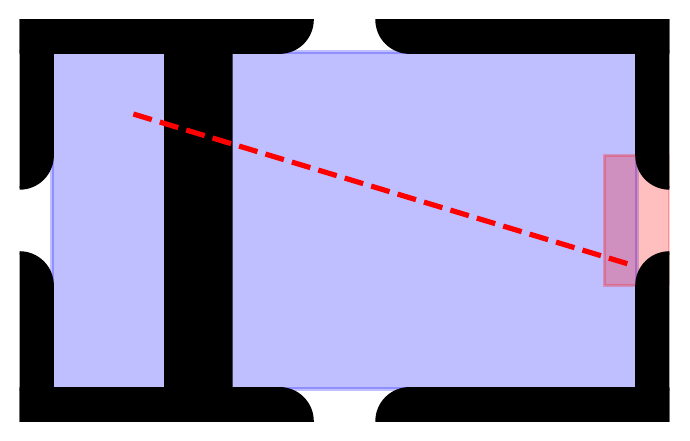}
    \caption{}
    \label{fig:convex:nonconvex}
  \end{subfigure}%
  \begin{subfigure}[t]{0.5\columnwidth}
    \centering
    \includegraphics[width=\textwidth]{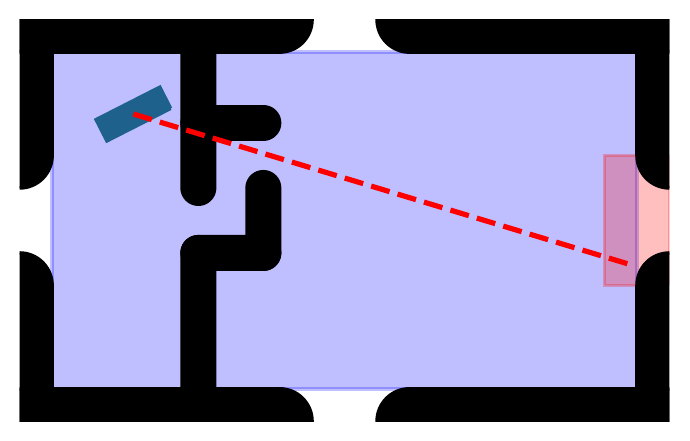}
    \caption{}
    \label{fig:convex:nonconvex_connected}
  \end{subfigure}
  \caption{%
    Useful regions for defining abstract operators are nearly convex.
    In all four examples here, the lower bound is given by the Euclidean distance in work space.
    In a convex region (a), the gap between the lower bound on the cost of a plan and the true optimal cost is zero.
    In an $\epsilon$-convex region (b), the gap between the lower bound $l$ on the cost of a plan and the true optimal cost $c$ is small: $l < c < (1+\epsilon) l$.
    Some regions are not $\epsilon$-convex for any finite $\epsilon$; for example, the region might not be connected (c).
    This can happen even if the region is connected in the work space (d) if it is not connected in configuration space.
    Here, the object cannot fit through the narrow gap, and so the region is not $\epsilon$-convex.
    In the presence of dynamic constraints, regions can fail to be path-connected even if they are connected in the workspace.
    \label{fig:convex}
  }
\end{figure}

Convexity is quite a strong requirement.
In a cluttered environment, a convex representation may need to contain many regions.
We can relax the requirement of convexity, and generalize to costs besides path length, by defining $\epsilon$-convexity.
A region $R$ is $\epsilon$-convex if
\begin{equation}
  \underset{\PrimitivePlan \in \PrimitivePlans_R(x, x')}{\mathrm{inf}} \mathcal{C}[\PrimitivePlan] \le (1+\epsilon)\Vert x - x'\Vert.
\end{equation}
This is shown graphically in figure~\ref{fig:convex}.
Intuitively, a region is $\epsilon$-convex if the shortest path between any two points is only slightly longer than the distance between the points.
For example, if $\mathcal{X} \subset \Reals^n$, a convex region $R$ cluttered with convex objects of diameter less than $d$ is $\epsilon$-convex, with $\epsilon = \pi \sqrt{\frac{n}{2(n+1)}}$; this is an elementary consequence of Jung's theorem \cite{jung1899kleinste}.

\subsection{An Abstraction for the Door Puzzle}
\label{sec:abstraction:door_puzzle}

\begin{figure}
  \centering
    \begin{subfigure}[t]{\columnwidth}
      \centering
      \includegraphics[width=\textwidth]{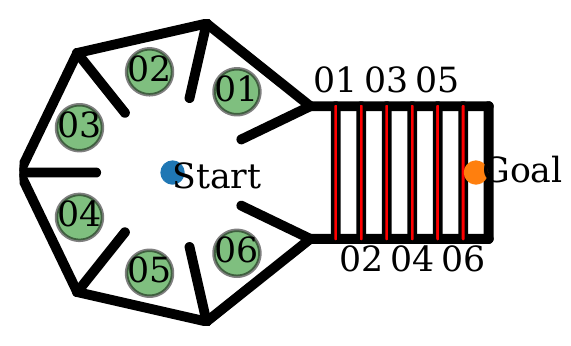}
      \caption{Problem}
      \label{fig:door_puzzle:tsp}
    \end{subfigure}
    \begin{subfigure}[t]{\columnwidth}
      \centering
      \includegraphics[width=\textwidth]{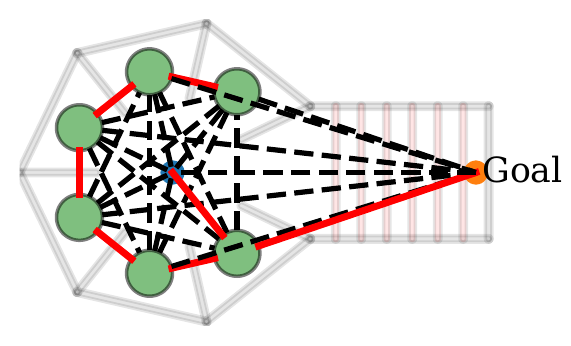}
      \caption{Lower bound}
      \label{fig:door_puzzle:mst}
    \end{subfigure}
    \begin{subfigure}[t]{\columnwidth}
      \centering
      \includegraphics[width=\textwidth]{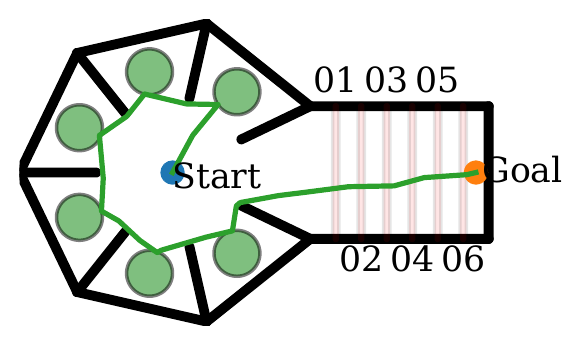}
      \caption{Solution}
      \label{fig:door_puzzle:plans}
    \end{subfigure}
  \caption{%
    In the problem shown in (a), it is easy to conclude that all $N=6$ doors must be opened before the robot can reach the goal.
    However, there are $N!=720$ possible orders in which we might press the switches.
    We can bound the cost of any sequence by solving a travelling salesperson problem (b, dotted lines), where the edge costs are the minimal distance the robot must travel to move between switches.
    Although this is an NP-hard problem, we can compute a lower bound on the cost of a solution in polynomial time by computing a minimum spanning tree (b, solid red line).
    This allows the planner to quickly find a near-optimal solution (c).
    \label{fig:door_puzzle}
  }
\end{figure}

The door puzzle introduced in the introduction combines the motion-planning aspects of navigation with a high-level task planning problem: the choice of which doors to open and in which order.
Unlike in the navigation problem, the configuration space for the door problem involves discrete components: $\mathcal{X} \subset \mathbb{R}^2 \times \{0, 1\}^N$, where where $N$ is the number of doors.
This creates an element of combinatorial search that is not present in the navigation experiment.

We use the same region-based abstraction to guide the search for motion plans, and construct a relaxed representation of the effects of toggling switches in PDDL by omitting geometric constraints like collision.
Using this representation, we can quickly compute a partial ordering on the sequence of switches that need to be pressed in order to reach the goal.
For example, in figure~\ref{fig:door_puzzle}, the path to the goal is blocked by six doors.
Before we can move towards the goal, we must move to and press each of the six switches.
This leaves us with the task of computing a lower bound on the cost to reach and toggle each switch.
We can find such a bound in two steps.
First, we construct a directed graph whose vertices are the possible effects of executing each operator, and whose edges have weights that lower bound the cost of executing each operator.
This graph, and the minimum spanning tree, are drawn in figure~\ref{fig:door_puzzle:mst}.
This reduces the problem of finding a lower bound to solving a travelling salesperson problem (TSP).
While solving a TSP requires exponential time, we can compute a lower bound on the cost of the optimal solution by computing a minimum spanning tree of the directed graph---and this is a computation that can be done in polynomial time with standard methods.
Although this bound neglects possible interactions between the operators, it is admissible; in fact, it is an admissible special case of the more general (and inadmissible) $h_{\mathrm{add}}$ heuristic \cite{haslum2000admissible}.
We can use this bound to guide the search for a more detailed motion plan (figure~\ref{fig:door_puzzle:plans}).

\section{Algorithms}
\label{sec:algorithms}

We now describe several algorithms that leverage an admissible angelic abstraction to search efficiently, even in high-dimensional continuous spaces.
Our algorithms are all derived from the angelic hierarchical A* algorithm developed by \cite{marthi2008angelic}.
We begin by reviewing this algorithm (section~\ref{sec:algorithms:angelic}), then discuss a subtle variation that dramatically improves efficiency in some common degenerate or nearly degenerate cases (section~\ref{sec:algorithms:acyclic}).
Finally, we discuss an extension that solves the approximately optimal planning problem, embracing the key insights of weighted A* (section~\ref{sec:approximate}).

\subsection{Angelic A*}
\label{sec:algorithms:angelic}

Angelic A* (algorithm~\ref{alg:angelic}) is a reformulation of the angelic hierarchical A* algorithm \cite{marthi2008angelic}.
This algorithm solves the optimal planning problem using a best-first forward search over abstract plans.

The primary data structure maintained by our algorithm is a tree.
Each node in the tree is a tuple $(\AbstractOperator, \AbstractPlan_{-}, \textsc{Base}(\AbstractPlan), \hat{V}[\AbstractPlan])$ representing a plan $\AbstractPlan = \AbstractPlan_{-} \circ \AbstractOperator$, where
\begin{itemize}
  \item $\AbstractOperator$ is an abstract operator,
  \item $\AbstractPlan_{-}$ is a pointer to the predecessor of the node,
  \item $\textsc{Base}(\AbstractPlan)$ is a pointer to the base plan, which is used in choosing refinements, and
  \item $\hat{V}[\AbstractPlan]$ is an admissible bound on the valuation of $\AbstractPlan$.
\end{itemize}
The root of the tree is the node $(\varnothing, \varnothing, \varnothing, \{(\{x_s\}, \{x_s\}, 0, 0)\}$, representing the start of any plan.

\begin{algorithm}
  \caption{Angelic A*\label{alg:angelic}}
  \begin{algorithmic}[1]
    \Function{Search}{abstraction $(\mathcal{S}, \mathcal{A}, \bar{\mathcal{R}}, \hat{V})$}
      \State $\mathrm{root} = (\varnothing, \varnothing, \varnothing, \{(x_{\mathrm{s}}, x_{\mathrm{s}}, 0, 0)\})$ \label{alg:angelic:root}
      \State $\AbstractPlan^* = \varnothing$
      \State $\textsc{Bound}(\varnothing) = \hat{V}[\TopLevelOperator]$
      \State $\AbstractPlan_0 = \textsc{Propagate}(\mathrm{root}, [\TopLevelOperator])$ \label{alg:angelic:initialize}
      \State $Q = \{\AbstractPlan_0\}$ \label{alg:angelic:enqueue}
      \While {$\vert Q \vert > 0$}
      \State $\AbstractPlan = \mathrm{arg\,min} \{ \hat{V}[\AbstractPlan](\{x_s\}, X_g): \AbstractPlan \in Q \}$ \label{alg:angelic:min}
      \If {$\textsc{Primitive}(\AbstractPlan^*)$ and $ \hat{V}_U[\AbstractPlan^*] \prec \hat{V}_L[\AbstractPlan]$} \label{alg:angelic:check}
          \State\Return $\AbstractPlan^*$ \label{alg:angelic:success}
        \Else
          \State $Q \gets Q \setminus \{\AbstractPlan\}$ \label{alg:angelic:begin_expand}
          \State $S \gets \textsc{Successors}(\AbstractPlan)$ 
          \For{$\AbstractPlan' \in S$}
            \If {$\hat{V}_U[\AbstractPlan'] < \hat{V}_U[\AbstractPlan^*]$}
              \State $\AbstractPlan^* \gets \AbstractPlan'$ \label{alg:angelic:store_best}
            \EndIf
          \EndFor
          \State $Q \gets Q \cup \{\AbstractPlan' \in S: \neg \hat{V}_U[\AbstractPlan^*] \prec \hat{V}_L[\AbstractPlan]\}$ \label{alg:angelic:end_expand}
        \EndIf
      \EndWhile
      \State\Return $\varnothing$
    \EndFunction
    \Function{Successors}{plan node $\AbstractPlan$}
      \label{alg:angelic:start_successors}
      \State $\textsc{Post}(\textsc{Base}(\AbstractPlan)) = \{\AbstractState': (\AbstractState, \AbstractState', l, u) \in \hat{V}[\textsc{Base}(\AbstractPlan)]\}$
      \State $\AbstractOperator = \textsc{Operator}(\textsc{Head}(\AbstractPlan))$
      \State $S = \varnothing$
      \For {$\AbstractPlan' : (\AbstractOperator, \AbstractPlan') \in \bar{\mathcal{R}}, \exists \AbstractState \in \textsc{Post}(\textsc{Base}(\AbstractPlan)):\textsc{Head}(\AbstractPlan') \cap \AbstractState \ne \varnothing$}
        \State $\AbstractPlan_{\mathrm{ref}} \gets \textsc{Propagate}(\textsc{Base}(\AbstractPlan), \AbstractPlan' \circ \textsc{Ext}(\AbstractPlan))$ \label{alg:angelic:call_propagate}
        \If {$\hat{V}_L[\AbstractPlan_{\mathrm{ref}}](x_s, X_g) < \infty$}
          \State $S \gets S \cup \{\AbstractPlan_{\mathrm{ref}}\}$
        \EndIf
      \EndFor
      \State\Return $S$
      \label{alg:angelic:end_successors}
    \EndFunction
    \Function{Propagate}{base node $\AbstractPlan$, list $\AbstractPlan_{\mathrm{ext}}$}
        \label{alg:angelic:propagate}
        \State $\mathbf{b} \gets \AbstractPlan$
        \While {$\AbstractPlan_{\mathrm{ext}}$ is not empty}
          \State $\AbstractOperator \gets \textsc{Pop}(\AbstractPlan_{\mathrm{ext}})$
          \If {$\AbstractOperator$ is more primitive than $\textsc{Operator}(\AbstractPlan)$}
            \State $\mathbf{b} \gets \AbstractPlan$
          \EndIf
          \State $\AbstractPlan \gets (\AbstractOperator, \AbstractPlan, \mathbf{b}, \hat{V}[\AbstractPlan \circ \AbstractOperator])$
          \If {$\hat{V}[\AbstractPlan] = \varnothing$}
            \Return
          \ElsIf {$\textsc{Bound}(\AbstractPlan_{\mathrm{ext}}) \prec \hat{V}_L[\AbstractPlan]$}
            \State \Return $\varnothing$ \label{alg:angelic:prune}
          \Else
            \State $\textsc{Bound}(\AbstractPlan_{\mathrm{ext}}) \gets \textsc{Bound}(\AbstractPlan_{\mathrm{ext}}) \cup \hat{V}[\AbstractPlan]$
            \label{alg:angelic:join}
          \EndIf
        \EndWhile
        \State\Return $A$
    \EndFunction
  \end{algorithmic}
\end{algorithm}

The main entry point for the algorithm is the $\textsc{Search}$ routine, which first constructs the root plan node (line~\ref{alg:angelic:root}) then computes an initial abstract plan that includes all possible primitive plans (line~\ref{alg:angelic:initialize}).
This abstract plan is then added to the plan queue (line~\ref{alg:angelic:enqueue}).

Then, as long as a plan remains on the queue, AA* repeatedly finds the abstract plan in $Q$ with the lowest lower bound (line~\ref{alg:angelic:min}).
If this plan is dominated by a previously discovered plan, then the algorithm returns successfully, as any remaining plan on the queue is also dominated.
Otherwise, AA* expands the active plan by computing its successors and adding them to the queue if they cannot be pruned (lines~\ref{alg:angelic:begin_expand}-\ref{alg:angelic:end_expand}).
If the queue becomes empty without discovering a primitive plan that reaches the goal, then no plan exists and the algorithm returns failure.

\begin{figure*}
  \centering
    \begin{subfigure}[t]{\textwidth}
      \centering
      \begin{tikzpicture}[scale=1, every label/.append style = {font=\tiny} ]
  \tikzset{>=latex}
  \tikzstyle{state}=[rectangle,inner sep=0.05cm,minimum size=0.75cm,node distance=1.8cm]
  \node[circle,fill=black,inner sep=0pt,minimum size=3pt, label={$[0, 0]$}] (root) {};
  \node[state,right=0.45cm of root, label={$[1.6,1.6]$}] (a) {$(-6.3, 5.9)$};
  \node[state,right=0.35cm of a, label={$[3.3,4.0]$}] (A) {$\textsc{Go}(S_1)$};
  \node[state,right=0.35cm of A, label={$[4.3,5.0]$}] (B) {$\textsc{Toggle}(S_1)$};
  \node[state,right=0.35cm of B, label={$[7.3,9.1]$}] (C) {$\textsc{Go}(R_1, R_2)$};
  \node[state,right=0.35cm of C, label={$[11.2,15.0]$}] (D) {$\textsc{Go}(R_2, X_g)$};
  \path[<-]
    (root) edge (a)
    (a) edge (A)
    (A) edge (B)
    (B) edge (C)
    (C) edge (D)
    ;

  \draw [decorate,decoration={brace,amplitude=0.2cm,raise=0.3cm,mirror}] (root.west) -- node [midway,below=0.4cm] {$\textsc{Base}(\mathbf{p})$} (a.east);
  \draw [decorate,decoration={brace,amplitude=0.2cm,raise=0.3cm,mirror}] (A.west) -- node [midway,below=0.4cm] {$\textsc{Head}(\mathbf{p})$} (A.east);
  \draw [decorate,decoration={brace,amplitude=0.2cm,raise=0.3cm,mirror}] (B.west) -- node [midway,below=0.4cm] {$\textsc{Ext}(\mathbf{p})$} (D.east);
\end{tikzpicture}
    \end{subfigure}
    \begin{subfigure}[t]{\textwidth}
      \centering
      \begin{tikzpicture}[scale=1, every label/.append style = {font=\tiny}]
  \tikzset{>=latex}
  \tikzstyle{state}=[rectangle,inner sep=0.05cm,minimum size=0.75cm,node distance=2cm]
  \node[circle,fill=black,inner sep=0pt,minimum size=3pt, label={$[0, 0]$}] (root) {};
  \node[state,right=0.45cm of root, label={$[1.6,1.6]$}] (a) {$(-6.3, 5.9)$};
  \node[state,right=0.45cm of a] (b) {\color{red} $(-7.2, 7.1)$};
  \node[state,right=0.35cm of b] (A) {\color{red} $\textsc{Go}(S_1)$};
  \node[state,right=0.35cm of A] (B) {$\textsc{Toggle}(S_1)$};
  \node[state,right=0.35cm of B] (C) {$\textsc{Go}(R_1, R_2)$};
  \node[state,right=0.35cm of C] (D) {$\textsc{Go}(R_2, X_g)$};
  \path[<-]
    (root) edge (a)
    ;
  \path[<-,color=red]
    (b) edge (A)
    ;
  \path[<-]
    (B) edge (C)
    (C) edge (D)
    ;

  \draw [decorate,decoration={brace,amplitude=0.2cm,raise=0.3cm,mirror}] (root.west) -- node [midway,below=0.4cm] {$\textsc{Base}(\mathbf{p})$} (a.east);
  \draw [decorate,decoration={brace,amplitude=0.2cm,raise=0.3cm,mirror}] (b.west) -- node [midway,below=0.4cm] {$\mathbf{p}'\in\textsc{Ref}(\textsc{Head}(\mathbf{p}))$} (A.east);
  \draw [decorate,decoration={brace,amplitude=0.2cm,raise=0.3cm,mirror}] (B.west) -- node [midway,below=0.4cm] {$\textsc{Ext}(\mathbf{p})$} (D.east);
\end{tikzpicture}
    \end{subfigure}
    \begin{subfigure}[t]{\textwidth}
      \centering
      \begin{tikzpicture}[scale=1, every label/.append style = {font=\tiny}]
  \tikzset{>=latex}
  \tikzstyle{state}=[rectangle,inner sep=0.05cm,minimum size=0.75cm,node distance=2cm]
  \node[circle,fill=black,inner sep=0pt,minimum size=3pt, label={$[0, 0]$}] (root) {};
  \node[state,right=0.45cm of root, label={$[1.6,1.6]$}] (a) {$(-6.3, 5.9)$};
  \node[state,right=0.45cm of a, label={$[3.1,3.1]$}] (b) {$(-7.2, 7.1)$};
  \node[state,right=0.35cm of b, label={$[3.5,4.2]$}] (A) {$\textsc{Go}(S_1)$};
  \node[state,right=0.35cm of A, label={$[4.5,5.2]$}] (B) {$\textsc{Toggle}(S_1)$};
  \node[state,right=0.35cm of B, label={$[7.5,9.3]$}] (C) {$\textsc{Go}(R_1, R_2)$};
  \node[state,right=0.35cm of C, label={$[11.4,15.2]$}] (D) {$\textsc{Go}(R_2, X_g)$};
  \path[<-]
    (root) edge (a)
    (a) edge (b)
    (b) edge (A)
    (A) edge (B)
    (B) edge (C)
    (C) edge (D)
    ;

  \draw [decorate,decoration={brace,amplitude=0.2cm,raise=0.3cm,mirror}] (root.west) -- node [midway,below=0.4cm] {$\textsc{Base}(\mathbf{p}_{\mathrm{ref}})$} (b.east);
  \draw [decorate,decoration={brace,amplitude=0.2cm,raise=0.3cm,mirror}] (A.west) -- node [midway,below=0.4cm] {$\textsc{Head}(\mathbf{p}_{\mathrm{ref}})$} (A.east);
  \draw [decorate,decoration={brace,amplitude=0.2cm,raise=0.3cm,mirror}] (B.west) -- node [midway,below=0.4cm] {$\textsc{Ext}(\mathbf{p}_{\mathrm{ref}})$} (D.east);
\end{tikzpicture}
    \end{subfigure}
  \caption{%
    A schematic illustration of the process by which we construct successor plans.
    A plan is represented by a collection of nodes representing operators.
    Each node has a pointer to its predecessor, and represents the concatenation of the predecessor with its operator.
    Each node also has a pointer to a \emph{base} node.
    To form the successors of a plan, we first break plan into three pieces: the base, the node after the base (called the head), and the rest of the plan (called the extension).
    We then replace the head with a valid refinement, chosen to be optimistically feasible.
    Finally, we propagate, creating new nodes corresponding to the operators in the refinement and the extension.
    \label{fig:successors}
  }
\end{figure*}
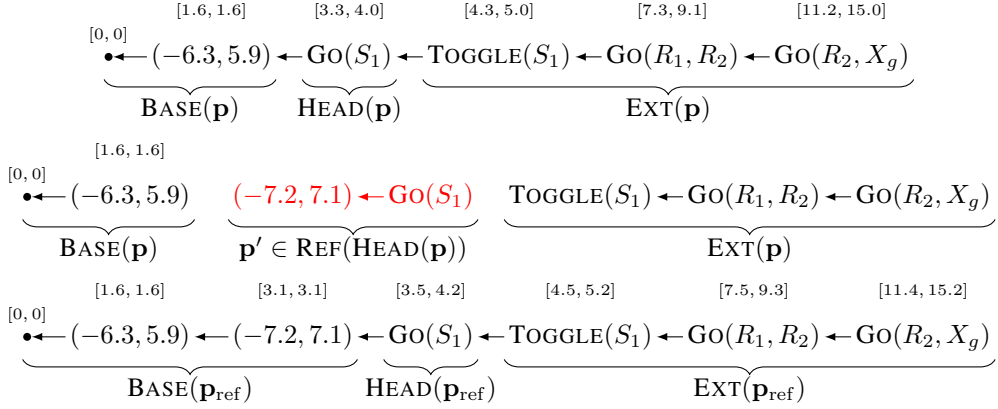

\begin{figure*}
  \centering
  \includegraphics[width=.8\textwidth]{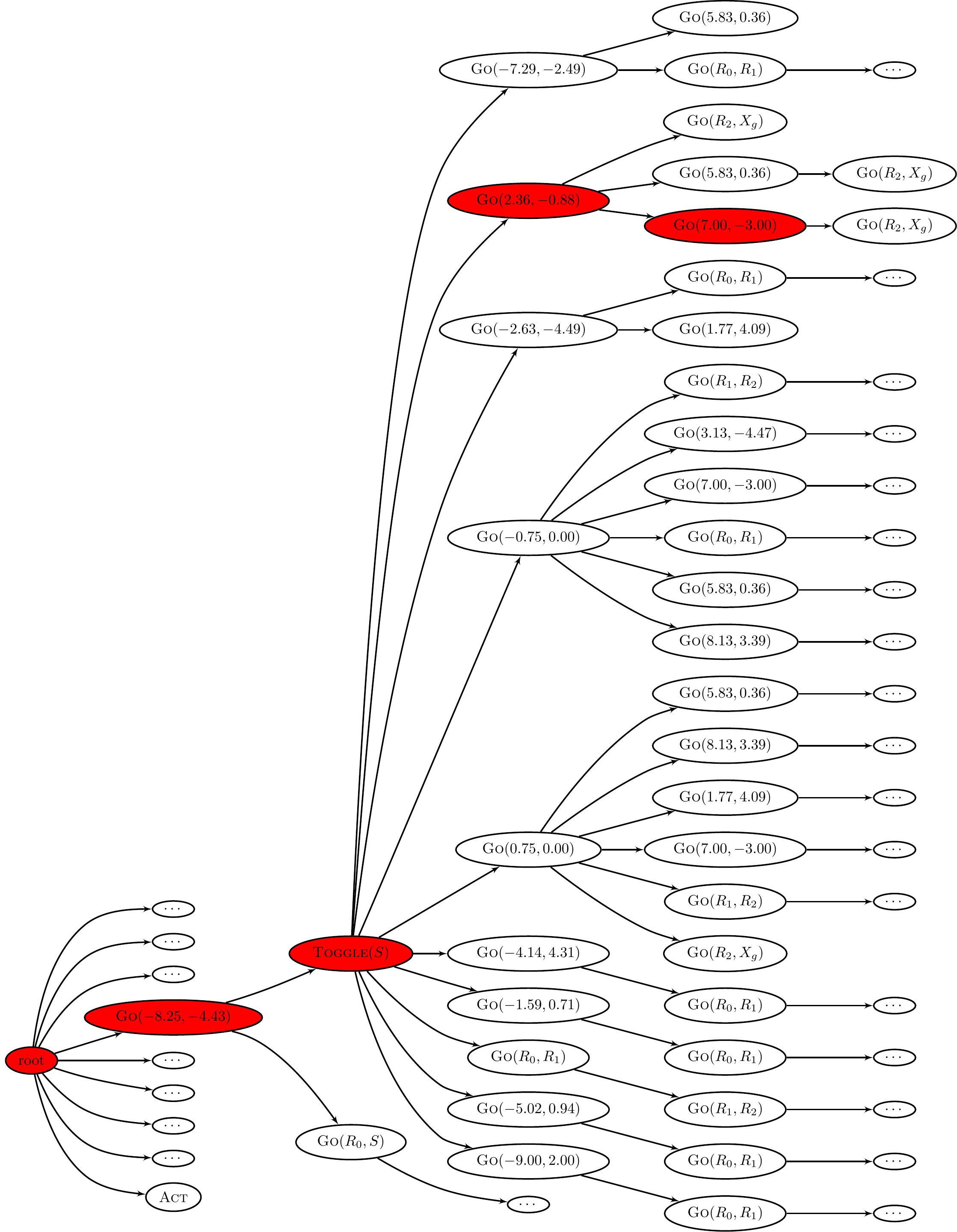}
  \caption{%
    Part of the plan tree constructed by AA* for the problem shown in figure~\ref{fig:plan_tree:problem}.
    Each node represents a plan; edges link a node to its predecessor.
    Nodes that are part of the optimal plan are highlighted in red.
    Branches of the tree not drawn are indicated with an ellipsis.
    The act of opening the door is referred to as $\textsc{Toggle}(S)$.
    Primitive motion operators are referred to as $\textsc{Go}(x, y)$, where $x$ and $y$ are coordinates.
    An abstract motion to a region $R_j$ through a region $R_i$ is referred to as $\textsc{Go}(R_i, R_j)$.
    The top level operator is labelled $\textsc{Act}$.
    \label{fig:plan_tree:solution}
  }
\end{figure*}
\begin{figure}
  \centering
  \includegraphics[width=\columnwidth]{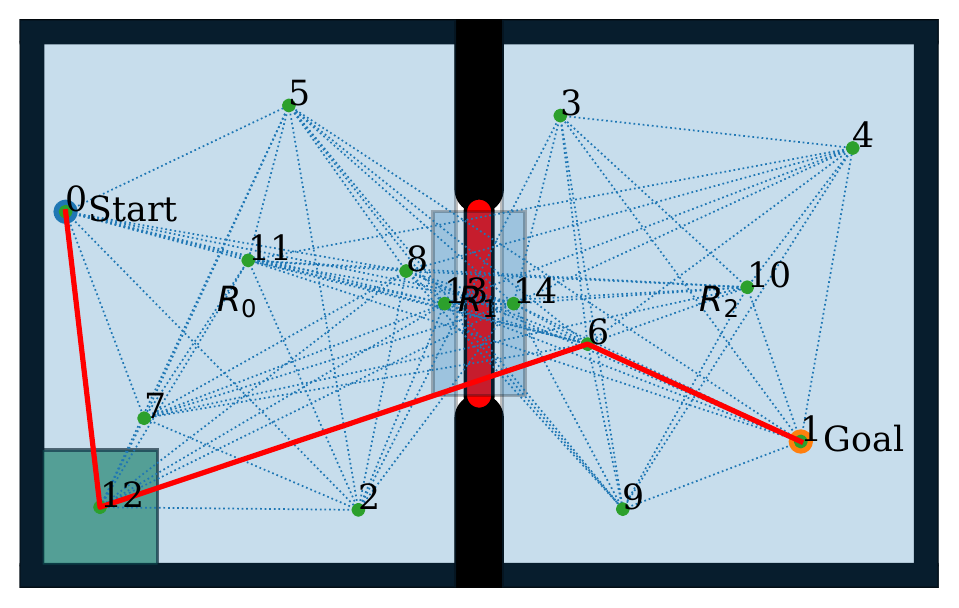}
  \caption{%
    The problem solved by the tree in figure~\ref{fig:plan_tree:solution}.
    The plan first toggles the switch in the lower left hand corner, then moves through the opened door to reach the goal.
    Primitive operators (edges) are displayed as dotted lines, while the optimal plan is highlighted in red.
    \label{fig:plan_tree:problem}
  }
\end{figure}

AA* generates successors to a plan using the refinement relation.
It then constructs a set of \emph{child} plans by selecting one operator from the plan and replacing it with its refinements.
Any successor plan that cannot possibly contain an acceptable solution is pruned, while any plan that could contain an acceptable solution is added to the priority queue.
The algorithm terminates when we remove a plan from the queue that is dominated by a previously expanded primitive plan.

We compute the valuation of each new plan incrementally (line~\ref{alg:angelic:propagate}).
If that new plan does not optimistically reach some state with lower cost than a previously explored plan ending with the same extension, we discard it (line~\ref{alg:angelic:prune}).
Otherwise, we update the bounds on any plan with the current extension to include the new plan (line~\ref{alg:angelic:join}).
Next, if the upper bound on the cost of reaching the goal under the new plan is better than any previous plan, we record this new plan as the best yet found (line~\ref{alg:angelic:store_best}).
Finally, if the lower bound on the cost of reaching the goal under the new plan is better than the upper bound under any previous plan, we add it to the set $Q$ of active plans.


Marthi~et~al.~showed this algorithm will return the optimal refinement of the top-level operator $\TopLevelOperator$ after a finite number of iterations, provided the lower bound on the cost of every operator is greater than zero.
\begin{theorem}
  \label{thm:angelic:hierarchical}
  Algorithm~\ref{alg:angelic} will return the optimal primitive refinement of the abstract plan $\TopLevelOperator$, provided the lower bound on the cost of every operator is strictly positive \cite{marthi2008angelic}.
\end{theorem}

However, if the abstraction is admissible, we can prove the following stronger claim.
\begin{restatable*}{theorem}{AngelicOptimality}
  \label{thm:angelic:optimality}
  If the abstraction $\mathcal{A}$ is admissible and a feasible plan exists, then algorithm~\ref{alg:angelic} returns an optimal sequence of primitive operators in finite time, provided the lower bound on the cost of every operator is greater than zero.
\end{restatable*}
\begin{corollary}
  If the set of primitive operators $\mathcal{A}_{0,n}$ is asymptotically optimal (equation~\ref{eq:asymptotically_optimal}), then
  \begin{equation}
    \lim_{n\to\infty}\mathrm{Pr}(C[\textsc{Search}(\mathcal{A}_n)] < (1 +\epsilon) c^*) = 1.
  \end{equation}
\end{corollary}
\begin{proof}
  See appendix~\ref{sec:proofs}, theorem~\ref{thm:angelic:optimality}.
\end{proof}

The distinction between these claims is subtle, but important.
Theorem~\ref{thm:angelic:hierarchical} implies hierarchical optimality: if a plan is returned, no better plan can be expressed as a refinement of the top-level operator.
Theorem~\ref{thm:angelic:optimality} implies primitive optimality: if a plan is returned, no better plan exists.
If we can ensure our abstraction is admissible, then using our abstraction provides the same guarantees as a direct search over the space of primitive plans, but may be much faster.

\subsection{Acyclic Angelic A*}
\label{sec:algorithms:acyclic}

\begin{figure}
  \centering
  \begin{subfigure}[t]{0.5\columnwidth}
    \centering
    \includegraphics[width=\textwidth]{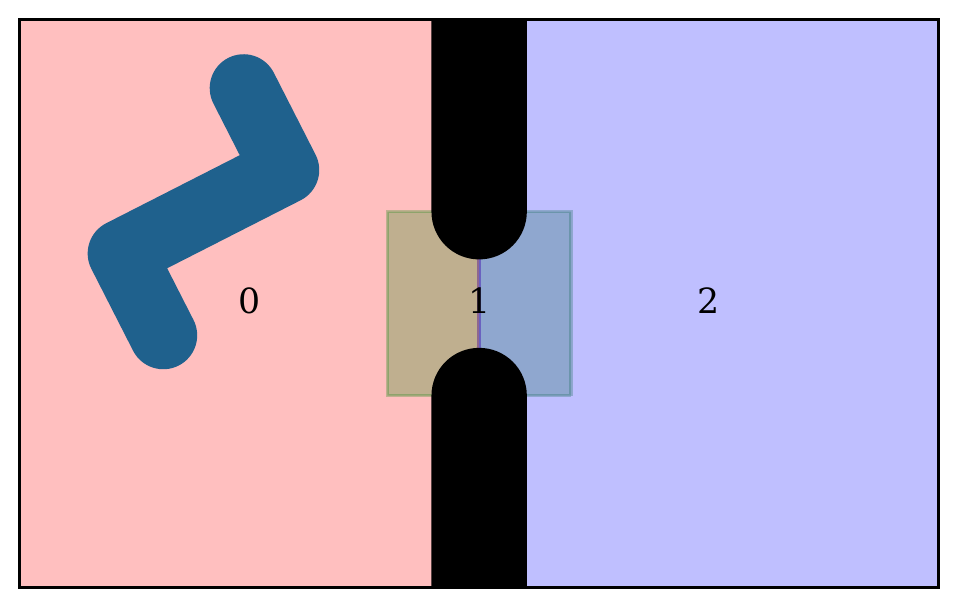}
    \caption{}
    \label{fig:acyclic:loop_zero}
  \end{subfigure}%
  \begin{subfigure}[t]{0.5\columnwidth}
    \centering
    \includegraphics[width=\textwidth]{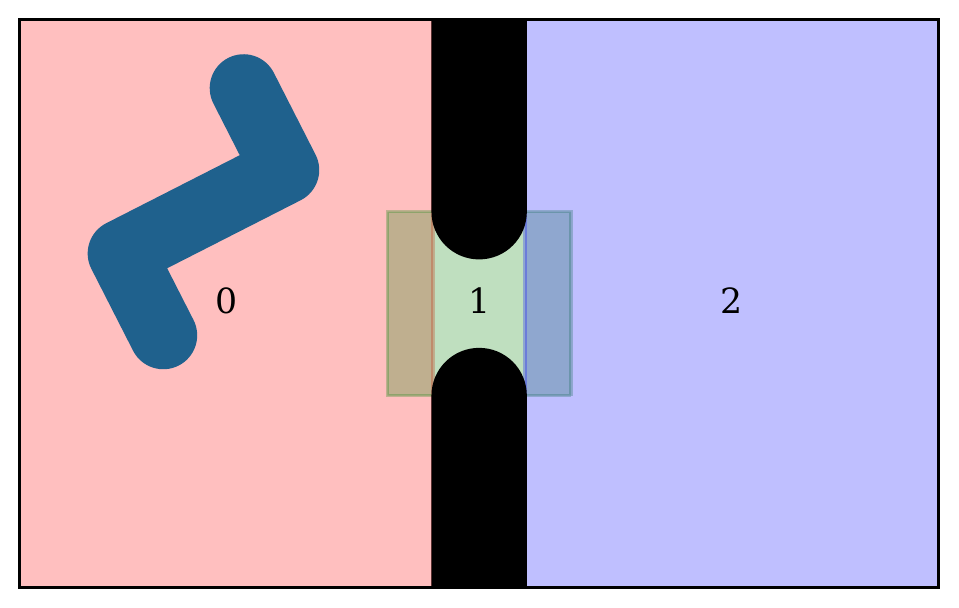}
    \caption{}
    \label{fig:acyclic:loop_small}
  \end{subfigure}
  \begin{subfigure}[t]{0.33\columnwidth}
    \centering
    \includegraphics[width=\textwidth]{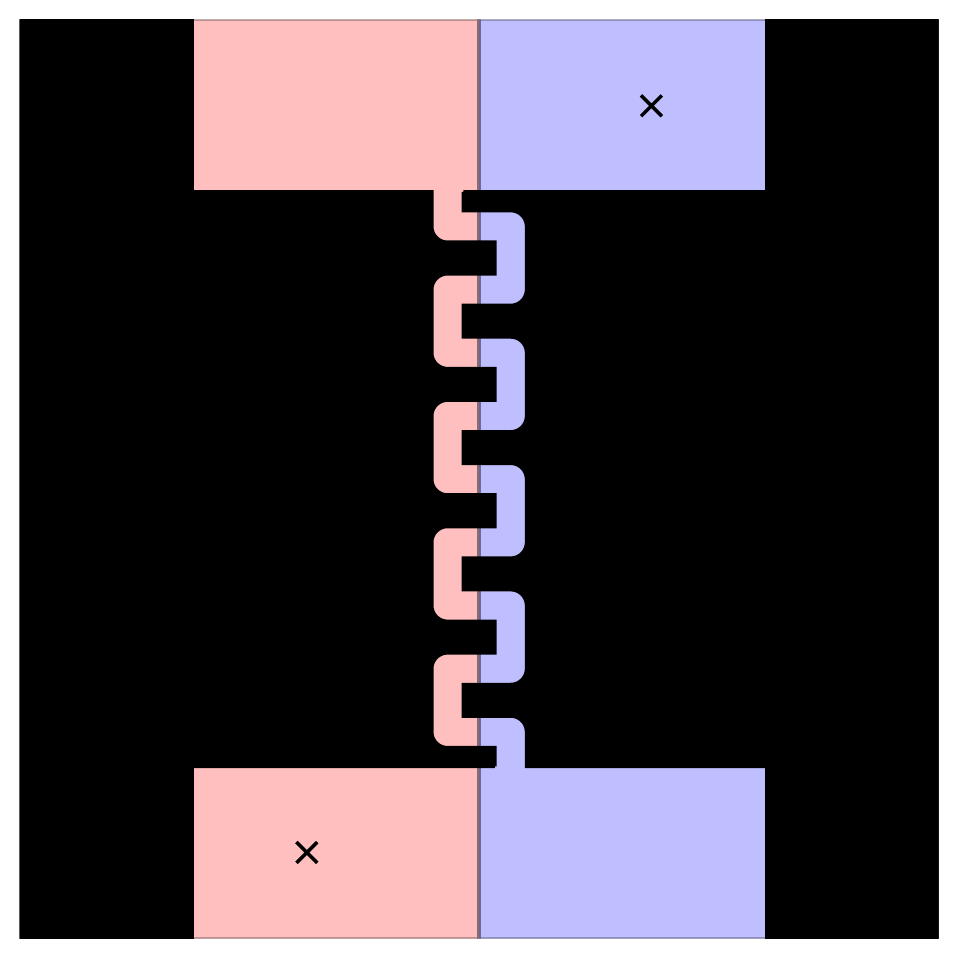}
    \caption{}
    \label{fig:acyclic:loop_necessary}
  \end{subfigure}%
  \begin{subfigure}[t]{0.33\columnwidth}
    \centering
    \includegraphics[width=\textwidth]{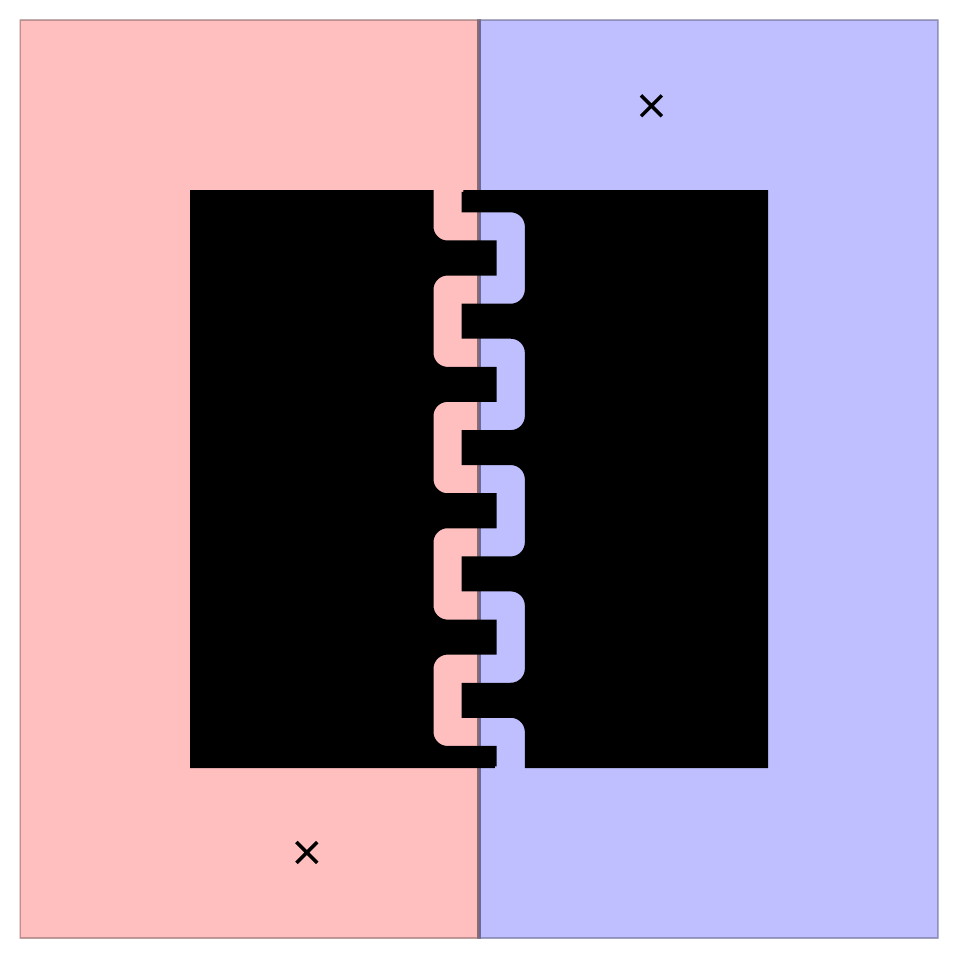}
    \caption{}
    \label{fig:acyclic:loop_expensive}
  \end{subfigure}%
  \begin{subfigure}[t]{0.33\columnwidth}
    \centering
    \includegraphics[width=\textwidth]{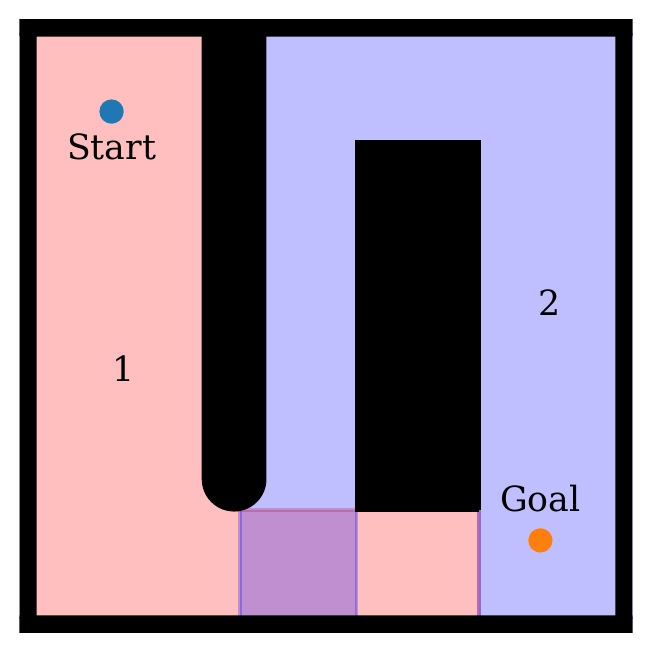}
    \caption{}
    \label{fig:acyclic:loop_connected}
  \end{subfigure}
  \caption{%
    An illustration of the problems with cyclic paths.
    Many natural operators in continuous domains have a cost with a lower bound of zero and no upper bound.
    For example, deciding whether the irregularly-shaped object (a) can reach the blue region requires detailed geometric analysis.
    Since regions $0$ and $2$ touch, the greatest lower bound on the cost of a plan in $\AbstractOperator_{01} \circ \AbstractOperator_{12}$ is the same as the bound on $\AbstractOperator_{01} \circ \AbstractOperator_{12} \circ \AbstractOperator_{10} \circ \AbstractOperator_{12}$.
    Any number of repetitions of the cycle $\AbstractOperator_{10} \circ \AbstractOperator_{12}$ will have the same cost, and so if $\AbstractOperator_{01} \circ \AbstractOperator_{12}$ is ever selected for expansion, the algorithm will only ever refine this infinite sequence of cyclic plans.
    Separating the regions (b) eliminates the infinite recursion, but remains inefficient; each cycle of $\AbstractOperator_{10} \circ \AbstractOperator_{12}$ adds only a small cost $\epsilon$ to the lower bound, meaning that if the next plan on the queue has a cost $\delta$ greater, the planner will consider $\lceil \delta / \epsilon \rceil$ cycles before considering the next acyclic plan.
    We cannot simply ignore these `cyclic' plans; in some scenarios, the best plan (c) or any feasible plan (d) is cyclic.
    This can occur even if the regions defining our operators are connected in configuration space: in the diagram in (e), although there is a feasible plan in $\AbstractOperator_{12} \circ \AbstractOperator_{2g}$, the optimal plan is in $\AbstractOperator_{12} \circ \AbstractOperator_{21} \circ \AbstractOperator_{2g}$.
    \label{fig:acyclic}
  }
\end{figure}

Algorithm~\ref{alg:angelic} requires strictly positive lower bounds on the cost of any operator.
In discrete problems, this is reasonable restriction, but it presents challenges in continuous problems.
For example, suppose we have a plan consisting of two operators $\AbstractOperator_{ij} \circ \AbstractOperator_{i'j'}$ from our navigation abstraction.
If the destination regions intersect---if $R_j \cap R_j' \ne \varnothing$---then the largest possible lower bound for the valuation of $\AbstractOperator_{i',j'}$ is zero.
This phenomenon can lead to a zero-cost cycle: a sequence of operators that can optimistically returns to a given state with zero cost (figure~\ref{fig:acyclic:loop_zero}).
Even positive cost-cycles are problematic, if the lower bound $l$ on the cost of a cycle is much smaller than the upper bound $u$: the algorithm can only prune a plan if it executes the cycle $\lceil u/l \rceil$ times (figure~\ref{fig:acyclic:loop_small}).
Unfortunately, we cannot simply discard any abstract plan with a cycle: the optimal plan may leave and return to an abstract state if the state is non-convex, even if the state is connected (figure~\ref{fig:acyclic:loop_necessary}-\ref{fig:acyclic:loop_connected}).
Often, this indicates a poor choice of abstraction, but it arise even with natural choices of abstraction, especially in domains with topologically complex configuration spaces.
We can deal with such edge cases while still avoiding cycles with a minor modification to the algorithm.

We define an acyclic plan as any plan $\AbstractPlan$ that cannot be partitioned into two plans $\AbstractPlan_0 \circ \AbstractPlan_1$ such that $\hat{V}_L[\AbstractPlan_0] \preceq \hat{V}_L[\AbstractPlan]$ (algorithm~\ref{alg:acyclic_fn}).
When we compute the successors of a plan $\AbstractPlan$, if we find the extension $\AbstractPlan_\mathrm{ext}$ would create a cyclic (i.e.~not acyclic) plan when propagated on top of $\textsc{Base}(\mathrm{\PrimitivePlan})$, we do not add $\AbstractPlan \circ \AbstractPlan_\mathrm{ext}$ to the set of successors.
Instead, we add $(\textsc{Base}(\AbstractPlan), \AbstractPlan_\mathrm{ext})$ to the set of deferred plans (algorithm~\ref{alg:acyclic}, line~\ref{alg:acyclic:defer}).

When any descendent of $\AbstractPlan$ is expanded, we consider activating any deferred extension of $\AbstractPlan$ by propagating it on top of the descendent plan.
If the resulting plan is no longer cyclic, we add it to the set of successors  (line~\ref{alg:acyclic:activate}).
This ensures that only acyclic plans will ever be added to the queue of plans, while also ensuring all plans that are not pruned will eventually be considered.

\begin{algorithm}
  \caption{Acyclic angelic A*\label{alg:acyclic}}
  \begin{algorithmic}[1]
    \Function{Successors}{plan node $\AbstractPlan$}
      \State\Comment $D$ is a global variable, initially set to $\varnothing$.
      \label{alg:acyclic:start_successors}
      \State $\textsc{Post}(\textsc{Base}(\AbstractPlan)) = \{\AbstractState': (\AbstractState, \AbstractState', l, u) \in \hat{V}[\textsc{Base}(\AbstractPlan)]\}$
      \State $S = \varnothing$
      \State $\AbstractOperator = \textsc{Operator}(\textsc{Head}(\AbstractPlan))$
      \For {$\AbstractPlan' : (\AbstractOperator, \AbstractPlan') \in \bar{\mathcal{R}}, \exists \AbstractState \in \textsc{Post}(\textsc{Base}(\AbstractPlan)):\textsc{Head}(\AbstractPlan') \cap \AbstractState \ne \varnothing$}
        \State $\AbstractPlan_{\mathrm{ref}} \gets \textsc{Propagate}(\textsc{Base}(\AbstractPlan), \AbstractPlan' \circ \textsc{Ext}(\AbstractPlan))$ \label{alg:acyclic:call_propagate}
        \If {$\hat{V}_L[\AbstractPlan_{\mathrm{ref}}](x_s, X_g) < \infty$}
          \If {$\textsc{Acyclic}(\AbstractPlan_{\mathrm{ref}}, \varnothing)$}
            \State $S \gets S \cup \{\AbstractPlan_{\mathrm{ref}}\}$
          \Else
            \State $D \gets D \cup \{(\textsc{Base}(\AbstractPlan), \textsc{Ext}(\AbstractPlan_{\mathrm{ref}}))\}$
            \label{alg:acyclic:defer}
          \EndIf
        \EndIf
      \EndFor
      \State $\AbstractPlan_a \gets \AbstractPlan$
      \While {$\textsc{Base}(\textsc{Parent}(\AbstractPlan_a)) \ne \varnothing$}
        \State $\AbstractPlan_a \gets \textsc{Base}(\textsc{Parent}(\AbstractPlan_a))$
        \For {$\AbstractPlan_{\mathrm{ext}} : (\AbstractPlan_a, \AbstractPlan_{\mathrm{ext}}) \in D$}
          \State $\AbstractPlan_{\mathrm{ref}} \gets \textsc{Propagate}(\textsc{Base}(\AbstractPlan), \AbstractPlan_{\mathrm{ext}})$
          \If {$\textsc{Acyclic}(\AbstractPlan_{\mathrm{ref}}, \varnothing)$ and $\hat{V}_L[\AbstractPlan_{\mathrm{ref}}] < \infty$}
            \State $S \gets S \cup \{\AbstractPlan_{\mathrm{ref}}\}$
            \label{alg:acyclic:activate}
          \EndIf
        \EndFor
      \EndWhile
      \State\Return $S$
      \label{alg:acyclic:end_successors}
    \EndFunction
  \end{algorithmic}
\end{algorithm}
\begin{algorithm}
  \caption{Acyclic angelic A*\label{alg:acyclic_fn}}
  \begin{algorithmic}[1]
    \Function{Acyclic}{plan nodes $\AbstractPlan, \AbstractPlan'$}
      \If {$\AbstractPlan = \varnothing$}
        \State\Return$\mathbf{true}$
      \ElsIf {$\AbstractPlan'=\varnothing$}
        \State $\AbstractPlan_- \gets \textsc{Predecessor}(\AbstractPlan)$
        \State\Return$\textsc{Acyclic}(\AbstractPlan_-, \varnothing) \wedge \textsc{Acyclic}(\AbstractPlan_-, \AbstractPlan)$
      \Else
        \State $\AbstractPlan_- \gets \textsc{Predecessor}(\AbstractPlan)$
        \State\Return$\neg (\hat{V}_L[\AbstractPlan] \preceq \hat{V}_L[\AbstractPlan']) \wedge \textsc{Acyclic}(\AbstractPlan_-, \varnothing)$
      \EndIf
    \EndFunction
  \end{algorithmic}
\end{algorithm}

\begin{restatable*}{theorem}{AcyclicOptimality}
  \label{thm:acyclic:optimality}
  If the abstraction $\mathcal{A}$ is admissible and a feasible plan exists, then the acyclic angelic A* algorithm returns a sequence of primitive operators with cost no greater than $c^*(\{x_s\}, X_g)$ in finite time.
\end{restatable*}
\begin{corollary}
  \label{thm:acyclic:asymptotic}
  If $\mathcal{A}_{0,n}$ is asymptotically optimal, then
  \begin{equation}
    \forall \epsilon \ge 0, \lim_{n\to\infty}\mathrm{Pr}(C[\textsc{Search}(\mathcal{A}_n)] < (1 +\epsilon) c^*) = 1.
  \end{equation}
\end{corollary}
\begin{proof}
  See appendix~\ref{sec:proofs}, theorem~\ref{thm:acyclic:optimality}.
\end{proof}

\subsection{Approximate Angelic A*}
\label{sec:approximate}

Even with a good abstraction, finding an optimal solution may be intractable for many problems.
By modifying the order in which plans are expanded and the conditions under which the algorithm terminates, we can accelerates the search process while still ensuring approximate optimality.
This modification is described in algorithm~\ref{alg:approximate}.

Often, admissible valuations are unduly optimistic: the lower bound $L[\AbstractPlan]$ on the cost of a plan is much less than the true optimal cost $c^*(\{x_s\}, X_g)$.
This problem is well-understood in the context of graph search, where it is often mitigated by using an admissible heuristic that has been inflated, as in weighted A* (\cite{pohl1970heuristic}).
WA* keeps a queue of states, and expands the state minimizing
\begin{equation}
  \textsc{Key}_{\mathrm{WA}^*}(x; w) = g(x) + w h(x),
\end{equation}
where $g(x)$ is the estimated cost to reach a state $x$ and $h(x)$ is an admissible estimate of the cost to reach the goal from a state $x$.
This biases the search towards plans that pessimistically reach states close to the goal.
If $h(x)-c^*(\{x\}, X_g)$ has only shallow local minima, this will explore far fewer states before finding a path to the goal than A* would.
Moreover, when a path to the goal is found, it will have cost less than $w c^*(\{x_s\}, X_g)$.

Unfortunately, we cannot directly apply this computation of an inflated priority in the context of angelic search.
When we use angelic semantics, we may not have a distinct cost $g(x)$ to reach a state; we only have bounds on the cost of plans.
In order to apply the idea of WA* to angelic search, we need to compute a priority that satisfies the same properties as $\textsc{Key}_{\mathrm{WA}^*}(x; w)$.
A na\"ive approach, such as inflating the lower bound on each operator, does not have the desired effect: it would inflate the priority of all plans equally and would not affect the order in which plans are expanded.

A more reasonable approach might be to inflate the lower bounds on each nonprimitive operator, which would exactly equal the priority computed by WA* for a flat abstraction---but this approach does not properly take upper bounds into account.
Consider an operator $\AbstractOperator$ for which $\hat{V}_U[\AbstractOperator](\AbstractState, \AbstractState') = (1+\varepsilon) L[\AbstractOperator](\AbstractState, \AbstractState')$ for some abstract state pair $(\AbstractState, \AbstractState')$.
If $\varepsilon$ is zero, the operator would be treated as primitive and its lower bound would not be inflated.
If it is small but positive, it would be treated as nonprimitive.
This would artificially bias the search away from operators that are almost, but not quite, primitive.

To avoid this undesirable bias, we recursively compute a priority $\textsc{Key}(\AbstractPlan, w)$.
If $\AbstractPlan_{-}=\textsc{Predecessor}(\AbstractPlan)$,
\begin{equation}
  \textsc{Key}(\AbstractPlan; w) =
    \min(\textsc{Key}(\AbstractPlan_{-}) + w (\hat{V}_L[\AbstractPlan] - \hat{V}_L[\AbstractPlan_{-}]), \hat{V}_U[\AbstractPlan]),
\end{equation}
with $\textsc{Key}(\varnothing; w) = 0$.
$\textsc{Key}(\AbstractPlan; w)$ has several useful properties.
For each plan $\AbstractPlan$, $\textsc{Key}(\AbstractPlan; w)$ is no greater than the upper bound $U[\AbstractPlan]$ or the inflated lower bound $w L[\AbstractPlan]$.
If $w=1$, then $\textsc{Key}(\AbstractPlan; w) = L[\AbstractPlan]$
For a primitive plan, $\textsc{Key}(\AbstractPlan; w)$ equals the cost of the plan.
In a flat abstraction, $\textsc{Key}(\AbstractPlan; w)$ is precisely equal to the cost estimate used by WA*.

We refer to this approach as \emph{approximate} angelic A*, and present pseudocode in algorithm~\ref{alg:approximate}.
The pseudocode is substantially similar to algorithm~\ref{alg:angelic}, and identical subroutines have been omitted.
Changes are highlighted in red.

\begin{algorithm}
  \caption{Approximate angelic A*\label{alg:approximate}}
  \begin{algorithmic}[1]
    \Function{Search}{abstraction $(\mathcal{S}, \mathcal{A}, \bar{\mathcal{R}}, \hat{V})$, \textcolor{red}{weight $w$}}
      \State $\mathrm{root} = (\varnothing, \varnothing, \varnothing, \{(x_{\mathrm{s}}, x_{\mathrm{s}}, 0, 0)\})$
      \State $\AbstractPlan^* = \varnothing$
      \State $\textsc{Bound}(\varnothing) = \hat{V}[\TopLevelOperator]$
      \State $\AbstractPlan_0 = \textsc{Propagate}(\mathrm{root}, [\TopLevelOperator])$
      \State $Q = \{\AbstractPlan_0\}$
      \While {$\vert Q \vert > 0$}
        \State $\AbstractPlan = \textcolor{red}{\mathrm{arg\,min} \{
          \textsc{Key}(\AbstractPlan, w): \AbstractPlan \in Q
        \}}$
        \If {$\textsc{Primitive}(\AbstractPlan^*)$ and $ \hat{V}_U[\AbstractPlan^*] \prec \hat{V}_L[\AbstractPlan]$}
          \State\Return $\AbstractPlan^*$
        \Else
          \State $Q \gets Q \setminus \{\AbstractPlan\}$
          \State $S \gets \textsc{Successors}(\AbstractPlan)$ 
          \For{$\AbstractPlan' \in S$}
            \If {$\hat{V}_U[\AbstractPlan'] < \hat{V}_U[\AbstractPlan^*]$}
              \State $\AbstractPlan^* \gets \AbstractPlan'$
            \EndIf
          \EndFor
          \State $Q \gets Q \cup S$
        \EndIf
      \EndWhile
      \State\Return $\varnothing$
    \EndFunction
  \end{algorithmic}
\end{algorithm}
\begin{algorithm}
  \caption{Approximate angelic search priority}
  \begin{algorithmic}[1]
    \Function{Key}{node $\AbstractPlan$, weight $w\in\mathbb{R}_{\ge 1}$}
      \If {$\AbstractPlan = \varnothing$}
        \State\Return $0$
      \Else
        \State $\AbstractPlan_- \gets \textsc{Predecessor}(\AbstractPlan)$
        \State\Return $\mathrm{min}(\textsc{Key}(\AbstractPlan_-) + w (\hat{V}_L[\AbstractPlan] - \hat{V}_L[\AbstractPlan_-]), \hat{V}_U[\AbstractPlan])$
      \EndIf
    \EndFunction
  \end{algorithmic}
\end{algorithm}

This algorithm is approximately optimal, in the sense that any plan it returns has cost less than $w$ times the cost of the optimal plan.
If we combine the acyclic generation of successor plans with the approximate search, the resulting algorithm is approximately optimal even if there are zero-cost operators.
\begin{restatable*}{theorem}{ApproximateOptimality}
  \label{thm:approximate:optimality}
  If the abstraction $\mathcal{A}$ is admissible and a feasible plan exists, then algorithm~\ref{alg:approximate} returns a sequence of primitive operators with cost no greater than $w \cdot c^*(\{x_s\}, X_g)$ in finite time.
\end{restatable*}
\begin{corollary}
  \label{thm:approximate:asymptotic}
  If $\mathcal{A}_{0,n}$ is asymptotically optimal, then
  \begin{equation}
    \forall \epsilon \ge 0, \lim_{n\to\infty}\mathrm{Pr}(C[\textsc{Search}(\mathcal{A}_n)] < (1 +\epsilon) w \cdot c^*) = 1.
  \end{equation}
\end{corollary}
\begin{proof}
  See appendix~\ref{sec:proofs}, theorem~\ref{thm:approximate:optimality}.
\end{proof}

If we use this idea in the angelic A* algorithm, rather than the acyclic angelic A* algorithm, the same theorem holds with the additional constraint that all operators must have a strictly positive lower-bound.


\section{Results}

We implemented algorithms~\ref{alg:angelic}-\ref{alg:approximate} and the abstractions described in sections~\ref{sec:abstraction:navigation} and \ref{sec:abstraction:door_puzzle} in the Python programming language.
We then compared the performance of the planner to the original angelic A* search algorithm (\cite{marthi2008angelic}) and to a search without abstraction using A*.

\begin{figure}
  \centering
    \begin{subfigure}[t]{0.5\columnwidth}
      \centering
      \includegraphics[width=\textwidth]{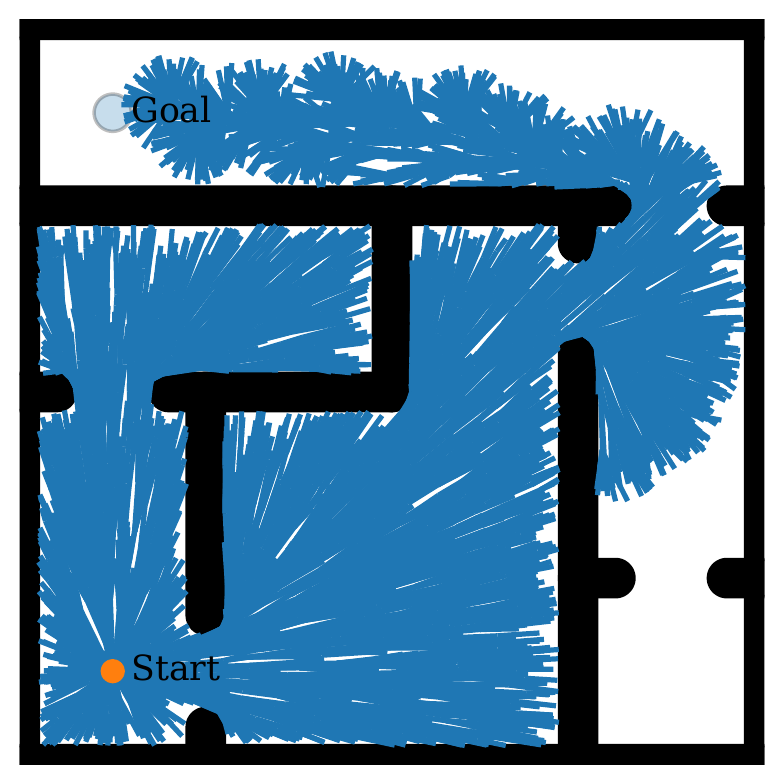}
      \caption{A*}
      \label{fig:navigation:astar}
    \end{subfigure}%
    \begin{subfigure}[t]{0.5\columnwidth}
      \centering
      \includegraphics[width=\textwidth]{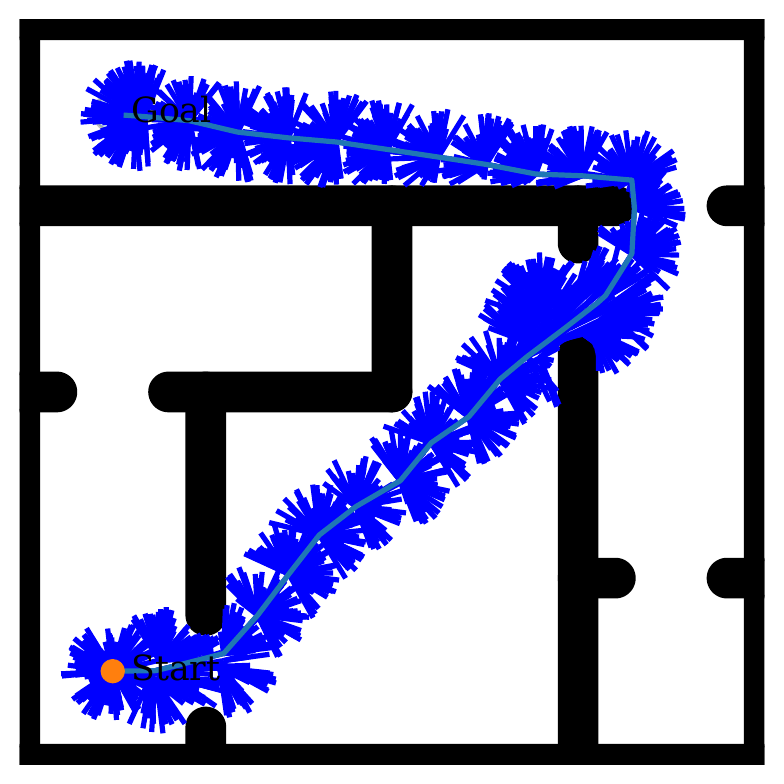}
      \caption{Acyclic Angelic A*}
      \label{fig:navigation:angelic}
    \end{subfigure}
  \caption{%
    The search trees constructed by A* (a) and by algorithm~\ref{alg:acyclic} (b).
    Note that the A* search needs to explore almost the entire space, due to limitations of the Euclidean distance as a heuristic.
    In contrast, when provided with a decomposition of the world into nearly-convex regions, angelic A* can find a path to the goal while exploring far fewer states.
    By avoiding plans with cycles, our modified angelic planning algorithm can explore these states while expanding far fewer plans.
    \label{fig:navigation}
  }
\end{figure}

In the navigation domain, we constructed a random discretization with $10^4$ states.
Examples of the search trees constructed by $A*$ and by algorithm~\ref{alg:acyclic} are given in figure~\ref{fig:navigation}.
By using the abstraction, the algorithm can avoid exploring large parts of the configuration space.
Our quantitative results bear this out: using abstraction allows us to reduce the number of states explored by a factor of three and the number of plans considered by several orders of magnitude.

\begin{figure}
  \centering
    \begin{subfigure}[t]{\columnwidth}
      \centering
      \includegraphics[width=\textwidth]{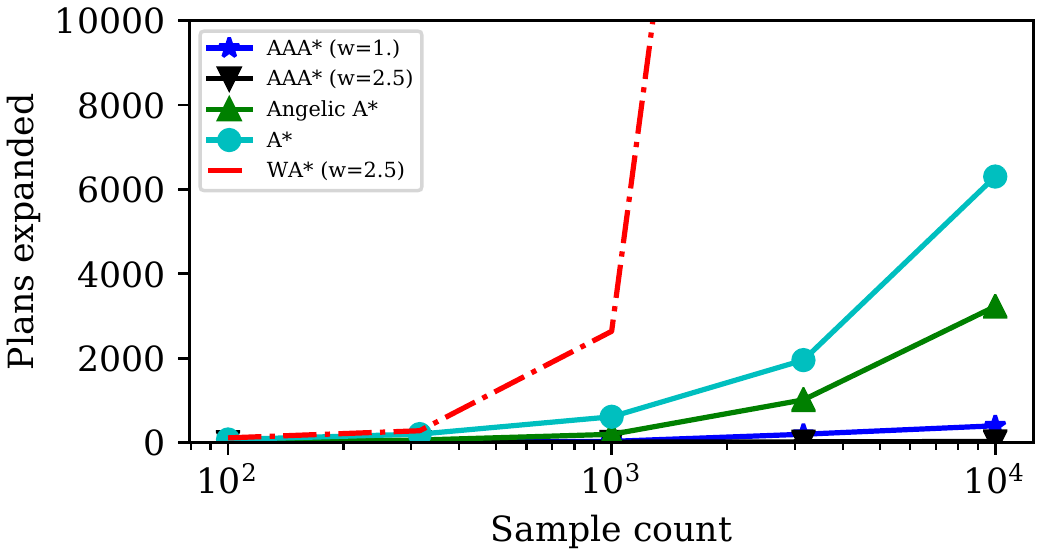}
    \end{subfigure}
    \begin{subfigure}[t]{\columnwidth}
      \centering
      \includegraphics[width=\textwidth]{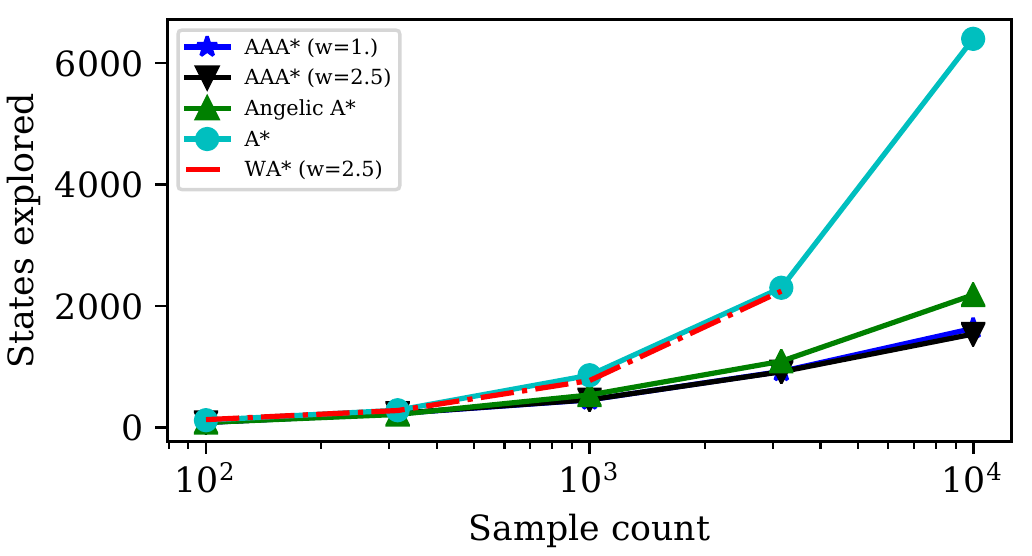}
    \end{subfigure}
  \caption{%
    Quantitative evaluation on an easy instance of the door puzzle domain with only two doors.
    More difficult instances could not be solved by any algorithm considered except algorithm~\ref{alg:acyclic}.
    The abscissa measures the number of randomly sampled states in the discretization of the configuration space.
    The ordinate axes measure the number plans expanded by each algorithm and the number of distinct configurations explored during search.
    \label{fig:quantitative}
  }
\end{figure}

Using abstraction in the door puzzle domain resulted in even larger speedups.
Even in easy problem instances with only a few doors, search without abstraction quickly became infeasible (figure~\ref{fig:quantitative}).
Using abstraction reduced the number of states explored by orders of magnitude.
However, the unmodified angelic search spent a great deal of time exploring plans with cycles.
By deferring these plans, our algorithms were able to reduce the number of plans expanded by an order of magnitude.
In fact, only our algorithm was able to solve problem instances with more than ten doors.
We were able to find 2-optimal plans for instances with up to 32 doors and $10^4$ sampled configurations (corresponding to a discretized state space with approximately 40 trillion states).
Unfortunately, software limitations prevented us from experimenting on states with more than 32 doors.

\begin{table}
  \footnotesize
  \centering
  \begin{tabular}{lrrrr}
\toprule
{} &    cost &     time &    plans &  states \\
\midrule
\textbf{A*          } &  33.430 &   42.119 &  11807 &  7948 \\
\textbf{Angelic A*  } &  33.430 &  160.256 &  25770 &  4758 \\
\textbf{AAA* (w=1.) } &  33.430 &    4.159 &    706 &  3068 \\
\textbf{AAA* (w=2.5)} &  35.586 &    0.697 &     48 &  1443 \\
\bottomrule
\end{tabular}

  \caption{Quantitative performance on a problem instance in the navigation domain.
    The discretized state space includes $10^4$ sampled configurations.
    We see that abstraction and approximation result expanding fewer plans and exploring fewer states, yielding a faster search and optimal or nearly optimal results.
  \label{table:quantitative}
}
\end{table}

\section{Related Work}

There is a long history of using abstraction to solve robotic planning problems \cite{nilsson1984shakey,lozano1987handey}, and although our formulation is different from most standard approaches to task or motion planning, many authors \cite{alami1990geometrical,simeon2004manipulation} have employed our underlying approach of searching for paths through a graph of configurations connected by feasible motion plans.
Practical algorithms often overcome the high computational cost of searching these planning graphs using clever heuristics.
For example, aSyMov \cite{cambon2009hybrid} and FFRob \cite{garrett2016ffrob} both employ the fast-forward heuristic, augmented with information derived from the geometric and kinematic computation.
Like these approaches, our work is built atop a heuristically-guided search; however, angelic semantics allow us to define upper bounds which can be used to prune away abstract plans, and allow for admissible hierarchies of arbitrary depth.

Our definition of abstract plans is closely related to the notion of ``plan skeletons'' considered by several authors \cite{erdogan2013planning,desilva2013towards,lozano2014constraint}.
Plan skeletons fix a sequence of operators but leave continuous parameters undefined.
There are many approaches to determining the feasibility of a given skeleton; for example, \cite{toussaint2015logic} uses continuous optimization techniques to search for optimal values of the real-valued variables.
\cite{lozano2014constraint} fix a discretization of the continuous variables then find feasible values by formulating and solving a constraint satisfaction problem.
\cite{lagriffoul2014efficiently} use linear programming to find valid values of the free variables or prove that none exist.
The primary difference between our approach and these plan skeletons is the choice of formalism.
By defining our abstract operators as implicitly defined \emph{sets} of primitive motion plans, we can reason about plans at varying levels of abstraction in a unified way, which is essential to the generality of our guarantees.

Another approach to task and motion planning represents geometric information in a way amenable to search using classical AI search techniques.
For example, \cite{dornhege2010integrating} model geometric information as predicates that can be resolved by solving motion planning problems during the task planning process.
More recently, \cite{ferrer-mestres2017combined} show that by fixing a discretization, in some domains all geometric information can be represented compactly in planning languages more expressive than PDDL, avoiding the need to make geometric queries during the planning process.
Other authors \cite{erdem2011combining,srivastava2013using,dantam2016incremental} use the task planner as a partial or approximate representation of the underlying geometric task, which can be improved during search.
For instance, \cite{erdem2011combining} use a high-level task planner to find an optimal task plan, then use a motion planner to attempt to find a kinematically feasible primitive solution to that task plan.
If no feasible solution exists, additional kinematic constraints are extracted from the motion planner and provided to the task planner, and the process is repeated.

Many authors have devised planning algorithms tailored to more specific task and motion planning domains.
For example, the problem of navigation among movable obstacles has long been of practical interest, and probabilistically complete solutions have been known since 2008 \cite{stilman2008planning,nieuwenhuisen2008effective}.
Planning for non-prehensile manipulation has been addressed by \cite{dogar2011framework} and by \cite{barry2013hierarchical}.
Our work could provide a new analytical tool with which to study these special classes of problems, and perhaps formulate new algorithms with stronger performance guarantees.

\section{Conclusions}

We have defined conditions on an abstraction that allow us to accelerate planning while preserving the ability to find an optimal or near-optimal solution to complex motion planning problems.
We motivate these conditions by deriving two admissible abstractions and showing they improve the efficiency of search without adversely affecting the quality of the resulting solutions.
We view this work as a proof of concept, demonstrating that a good abstraction can render optimal planning feasible even on large problems.
The classical planning community has developed several powerful families of admissible heuristics \cite{haslum2000admissible}; by reformulating these heuristics to employ angelic abstractions, we may be able to obtain optimal or near-optimal solutions to practical manipulation planning problems.

\section*{Acknowledgements}
This research was sponsored by Northrop Grumman and by the Robotics Collaborative Technology Alliance (RCTA) of the US Army.
Their support is gratefully acknowledged.

\bibliographystyle{named}
\bibliography{references}

\clearpage
\appendix

\section{Proofs}
\label{sec:proofs}

\begin{proposition}
  \label{thm:bounds}
  Define operator bounds
  \begin{align}
    V_L[\AbstractOperator](\AbstractState, \AbstractState') &= \inf \{ \inf \{ V[\AbstractPlan](s, s'): s' \in \AbstractState' \}: s \in \AbstractState \} \\
    V_U[\AbstractOperator](\AbstractState, \AbstractState') &= \sup \{ \inf \{ V[\AbstractPlan](s, s'): s' \in \AbstractState' \}: s \in \AbstractState \} \\
    \hat{V}_L[\AbstractOperator](\AbstractState, \AbstractState') &=
      \inf \{l: (\AbstractState_0, \AbstractState_1, l, u) \in \hat{V}[\AbstractOperator], \nonumber \\
      &\hspace{1cm} \AbstractState \cap \AbstractState_0 \ne \varnothing, \AbstractState' \cap \AbstractState_1 \ne \varnothing\} \\
    \hat{V}_U[\AbstractOperator](\AbstractState, \AbstractState') &=
      \inf \{u: (\AbstractState_0, \AbstractState_1, l, u) \in \hat{V}[\AbstractOperator], \nonumber\\
      &\hspace{1cm}\AbstractState \subseteq \AbstractState_0, \AbstractState' \subseteq \AbstractState_1\}.
\end{align}
  If $\hat{V}[\AbstractOperator]$ is admissible, then $\hat{V}_L[\AbstractOperator](\AbstractState, \AbstractState') \le V_L[\AbstractOperator](\AbstractState, \AbstractState')$ and $V_U[\AbstractOperator](\AbstractState, \AbstractState') \le \hat{V}_U[\AbstractOperator](\AbstractState, \AbstractState')$ for all abstract state pairs $\AbstractState, \AbstractState'$.
\end{proposition}
\begin{proof}
  \begin{align}
    \forall \PrimitiveState &\in \AbstractState, \PrimitiveState' \in \AbstractState', \nonumber\\
    &\exists (\AbstractState_0, \AbstractState_1, l, u) \in \hat{V}[\AbstractOperator]: l \le V[\AbstractOperator](\PrimitiveState,\PrimitiveState'), \PrimitiveState \in \AbstractState, \PrimitiveState' \in \AbstractState' \\
    &\PrimitiveState \in \AbstractState \wedge \PrimitiveState \in \AbstractState_0 \implies \AbstractState \cap \AbstractState_0 \ne \varnothing \\
    &\PrimitiveState' \in \AbstractState' \wedge \PrimitiveState' \in \AbstractState_1 \implies \AbstractState' \cap \AbstractState_1 \ne \varnothing \\
    &\hat{V}_L[\AbstractOperator](\{\PrimitiveState\}, \{\PrimitiveState'\}) < V[\AbstractOperator](\PrimitiveState, \PrimitiveState') \\
    \therefore\quad& \hat{V}_L[\AbstractOperator](\AbstractState, \AbstractState') \le V_L[\AbstractOperator](\AbstractState, \AbstractState') \\
    \forall \PrimitiveState &\in \AbstractState, \PrimitiveState' \in \AbstractState', \nonumber\\
    &\forall (\AbstractState_0, \AbstractState_1, l, u) \in \hat{V}[\AbstractOperator]: \PrimitiveState \in \AbstractState, \PrimitiveState' \in \AbstractState': u \ge V[\AbstractOperator](\PrimitiveState,\PrimitiveState') \\
    &\AbstractState \subseteq \AbstractState_0 \wedge \PrimitiveState \in \AbstractState \implies \PrimitiveState \in \AbstractState_0 \\
    &\AbstractState' \subseteq \AbstractState_1 \wedge \PrimitiveState' \in \AbstractState' \implies \AbstractState' \in \AbstractState_0 \\
    &\hat{V}_U[\AbstractOperator](\{\PrimitiveState\}, \{\PrimitiveState'\}) \ge V[\AbstractOperator](\PrimitiveState, \PrimitiveState') \\
    \therefore\quad& \hat{V}_U[\AbstractOperator](\AbstractState, \AbstractState') \ge V_U[\AbstractOperator](\AbstractState, \AbstractState')
  \end{align}
  Then $\hat{V}_L[\AbstractOperator](\AbstractState, \AbstractState') \le V_L[\AbstractOperator](\AbstractState, \AbstractState')$ and $V_U[\AbstractOperator](\AbstractState, \AbstractState') \le \hat{V}_U[\AbstractOperator](\AbstractState, \AbstractState')$ for all abstract state pairs $\AbstractState, \AbstractState'$.
\end{proof}

\begin{proposition}
  \label{thm:propagation}
  Suppose $\hat{V}[\AbstractOperator]$ and $\hat{V}[\AbstractOperator']$ are admissible.
  Then the valuation bound
  \begin{align}
    \hat{V}[\AbstractOperator \circ \AbstractOperator'] =
    &\{(\AbstractState, \AbstractState''', l+l', u+u'):
      (\AbstractState, \AbstractState', l, u) \in \hat{V}[\AbstractOperator],
      \nonumber \\&\quad
      (\AbstractState'', \AbstractState''', l', u') \in \hat{V}[\AbstractOperator'],
      \AbstractState' \subseteq \AbstractState''
    \} \,\cup \nonumber \\
    & \{(\AbstractState, \AbstractState''', l+l', u):
      (\AbstractState, \AbstractState', l, u') \in \hat{V}[\AbstractOperator],
      \nonumber \\&\quad
      (\AbstractState'', \AbstractState''', l', \infty) \in \hat{V}[\AbstractOperator'],
      \AbstractState' \cap \AbstractState'' \ne \varnothing
    \}
  \end{align}
  is admissible.
\end{proposition}
\begin{proof}
  Let $p \circ p'$ be a plan in $\AbstractOperator \circ \AbstractOperator'$, where $p \in \AbstractOperator$ and $p' \in \AbstractOperator'$.
  Then because $\hat{V}[\AbstractOperator]$ and $\hat{V}[\AbstractOperator']$ are admissible,
  \begin{align}
    \exists &(\AbstractState, \AbstractState', l, u) \in \hat{V}[\AbstractOperator]: p(0) \in \AbstractState, p(1) \in \AbstractState', l \le \mathcal{C}[p] \\
    \exists &(\AbstractState'', \AbstractState''', l', u') \in \hat{V}[\AbstractOperator']: p'(0) \in \AbstractState'', p'(1) \in \AbstractState''', l' \le \mathcal{C}[p']
  \end{align}
  Because $p \circ p'$ is continuous, $\AbstractState' \cap \AbstractState''$ intersect.
  \begin{equation}
    p(1) = p'(0) \implies \AbstractState' \cap \AbstractState'' \ne \varnothing 
  \end{equation}
  Because the cost function is additive, $l+l'$ is a lower bound on the cost of $p \circ p'$.
  \begin{equation}
      \mathcal{C}[p \circ p'] = \mathcal{C}[p] + \mathcal{C}[p'] \ge l + l' 
  \end{equation}
  By construction, it follows that $(\AbstractState, \AbstractState''', l + l', u + u') \in \hat{V}[\AbstractOperator \circ \AbstractOperator']$, and therefore
  \begin{multline}
    \therefore \forall \PrimitivePlan \circ \PrimitivePlan' \in \AbstractOperator \circ \AbstractOperator', \exists (\AbstractState, \AbstractState', l, u) \in \hat{V}[\AbstractOperator \circ \AbstractOperator']: \\
    p(0) \in \AbstractState, p(1) \in \AbstractState', l \le \mathcal{C}[p \circ p'].
  \end{multline}
  Thus $\hat{V}[\AbstractOperator \circ \AbstractOperator']$ is an admissible lower bound.

  Let $\PrimitiveState, \PrimitiveState, \PrimitiveState''$ be primitive states.
  \begin{flalign}
    V[\AbstractOperator \circ &\AbstractOperator'](\PrimitiveState, \PrimitiveState'') \nonumber\\
    &\le V[\AbstractOperator \circ \AbstractOperator'](\PrimitiveState, \PrimitiveState') + V[\AbstractOperator \circ \AbstractOperator'](\PrimitiveState', \PrimitiveState'') \\
    &\le V[\AbstractOperator](\PrimitiveState, \PrimitiveState') + V[\AbstractOperator'](\PrimitiveState', \PrimitiveState'') \\
    &\le V_U[\AbstractOperator](\AbstractState, \AbstractState') + V_U[\AbstractOperator'](\AbstractState'', \AbstractState''') \forall \AbstractState' \subseteq \AbstractState'' \\
    &\le \hat{V}_U[\AbstractOperator](\AbstractState, \AbstractState') + \hat{V}_U[\AbstractOperator'](\AbstractState'', \AbstractState''') \\
    &\qquad\qquad\forall \AbstractState' \subseteq \AbstractState'', \PrimitiveState \in \AbstractState, \PrimitiveState' \in \AbstractState', \PrimitiveState''\in \AbstractState''' \nonumber \\
    &\le u + u' \\
    &\qquad\forall (\AbstractState, \AbstractState', l, u) \in \hat{V}[\AbstractOperator]: \PrimitiveState \in \AbstractState, \PrimitiveState' \in \AbstractState', \nonumber \\
    &\qquad\forall (\AbstractState'', \AbstractState''', l', u') \in \hat{V}[\AbstractOperator']: \PrimitiveState' \in \AbstractState'', \PrimitiveState'' \in \AbstractState''' \nonumber \\
    &\le \inf \{u: (\AbstractState, \AbstractState''', l, u)  \in \hat{V}[\AbstractOperator  \circ  \AbstractOperator']\} \\
    &= \hat{V}_U[\AbstractOperator  \circ  \AbstractOperator'](\PrimitiveState, \PrimitiveState')
  \end{flalign}
  Thus $\hat{V}[\AbstractOperator \circ \AbstractOperator']$ is an admissible upper bound.
\end{proof}

\begin{proposition}
  \label{thm:join}
  If $\hat{V}[\AbstractPlan]$ and $\hat{V}[\AbstractPlan']$ are admissible, then $\hat{V}[\AbstractPlan \cup \AbstractPlan'] = \hat{V}[\AbstractPlan] \cup \hat{V}[\AbstractPlan']$ is admissible.
\end{proposition}
\begin{proof}
  Observe that 
  \begin{equation}
    V[\AbstractPlan \cup \AbstractPlan'](s, s') = \min(V[\AbstractPlan](s, s'), V[\AbstractPlan'](s, s')).
  \end{equation}
  Then
  \begin{align}
    \hat{V}_U[\AbstractPlan \cup \AbstractPlan'](\{s\}, \{s\}')
    &\ge \hat{V}_U[\AbstractPlan](\{s\}, \{s\}') \\
    &\ge V[\AbstractPlan](s, s') \\
    &\ge V[\AbstractPlan \cup \AbstractPlan'](s, s')
  \end{align}
  so $\hat{V}[\AbstractPlan \cup \AbstractPlan']$ is a valid upper bound.
  Note this argument applies to $\hat{V}_U[\AbstractPlan](\{s\}, \{s\}')$ as well.

  Then
  \begin{align}
    \hat{V}_L[\AbstractPlan \cup &\AbstractPlan'](\{s\}, \{s\}') \nonumber \\
    &= \min(\hat{V}_L[\AbstractPlan](\{s\}, \{s\}'), \hat{V}_L[\AbstractPlan'](\{s\}, \{s\}')) \\
    &\le \min(V[\AbstractPlan](s, s'), V[\AbstractPlan'](s, s')) \\
    &= V[\AbstractPlan \cup \AbstractPlan'](s, s')
  \end{align}
  so $\hat{V}[\AbstractPlan \cup \AbstractPlan']$ is a valid lower bound.
\end{proof}

\begin{proposition}
  \label{thm:regions_admissible}
  Let $\mathcal{S} = \{R_i\}$ be a decomposition of the state space $\mathcal{X}$ into $N$ regions, such that $\cup_{i \in [N]} R_i = \mathcal{X}$.
  The refinement relation
  \begin{equation}
    \begin{aligned}
      \bar{\mathcal{R}} =  \bigcup_{ij}
      & \{(\TopLevelOperator, \AbstractOperator_{ij} \circ \TopLevelOperator), (\TopLevelOperator, \AbstractOperator_{ij}) \} \,\cup \\
      & \{(\AbstractOperator_{ij}, \PrimitiveOperator \circ \AbstractOperator_{ij}): \PrimitiveOperator(t) \in R_i \forall t\} \,\cup \\
      & \{(\AbstractOperator_{ij}, \PrimitiveOperator ): \PrimitiveOperator(t) \in R_i \forall t, \PrimitiveOperator(1) \in \mathrm{cl}(R_j) \}
    \end{aligned}
  \end{equation}
  is admissible.
\end{proposition}
\begin{proof}
  Choose any $\PrimitivePlan \in \TopLevelOperator$.
  Because $\cup_{i \in [N]} R_i = \mathcal{X}$, there exists some $R_i$ such that $\PrimitivePlan(0) \in R_i$.
  If $\PrimitivePlan(t) \in R_i \forall t$, then $\PrimitivePlan \in \AbstractOperator{ii}$.
  Otherwise, there is some $t' \in [0, 1]$ such that $\PrimitivePlan(t') \in \partial R_i$ and $\PrimitivePlan(t) \in R_i \forall t < t'$.
  Because $\cup_{j \in [N]} R_j = \mathcal{X}$, there exists some $R_j$ such that $\PrimitivePlan(t') \in \mathrm{cl}(R_j)$.
  Then we can partition $\PrimitivePlan$ into $\PrimitivePlan_1 \circ \PrimitivePlan_2$, such that $\PrimitivePlan_1(t) \in R_i \forall t$ and $\PrimitivePlan_2(0) \in \mathrm{cl}(R_j)$.
  Then $\PrimitivePlan_1 \in \AbstractOperator_{ij}$ and $\PrimitivePlan_2 \in \TopLevelOperator$, so $\PrimitivePlan \in \AbstractOperator_{ij} \circ \TopLevelOperator$.

  Choose any $\PrimitivePlan \in \AbstractOperator_{ij}$.
  Because $\PrimitivePlan \in \PrimitiveOperators^*$, either we can partition $\PrimitivePlan$ into $\PrimitiveOperator \circ \PrimitivePlan'$, or $\PrimitivePlan = \PrimitiveOperator$.
  If $\PrimitivePlan = \PrimitiveOperator$, then $(\AbstractOperator_{ij}, \PrimitivePlan) \in \bar{\mathcal{R}}$.
  Otherwise, $\PrimitivePlan' \in \AbstractOperator_{ij}$, so $(\AbstractOperator_{ij}, \PrimitiveOperator \circ \PrimitivePlan') = (\AbstractOperator_{ij}, \PrimitivePlan) \in \bar{\mathcal{R}}$.
\end{proof}

\AngelicOptimality
\begin{proof}
  Define the condition $I_1(\PrimitivePlan)$ as
  \begin{equation}
    I_1(\PrimitivePlan) \iff \exists \AbstractPlan \in Q : \PrimitivePlan \in \AbstractPlan \vee \hat{V}_U[\AbstractPlan] \prec V[\PrimitivePlan].
  \end{equation}
  We will first show that if $I_1(\PrimitivePlan)$ holds before line~\ref{alg:angelic:begin_expand}, it holds after line~\ref{alg:angelic:end_expand}. 
  If $I_1(\PrimitivePlan)$ holds initially, there is a plan $\AbstractPlan \in Q$ such that $\PrimitivePlan \in \AbstractPlan$.
  If $\AbstractPlan \ne \mathrm{arg\,min} \{\hat{V}_L[\AbstractPlan]: \AbstractPlan\in Q\}$, then $\AbstractPlan \in Q$ after line~\ref{alg:angelic:end_expand}.
  Otherwise, because the abstraction is admissible, there is a plan $\AbstractPlan'$ such that $\PrimitivePlan \in \AbstractPlan'$ and $(\AbstractPlan, \AbstractPlan') \in \mathcal{R}$.
  If $P \prec \AbstractPlan'$, there is a plan $\AbstractPlan'' \in \mathbf{Q}$ such that $\hat{V}_U[\AbstractPlan''] \prec V[\PrimitivePlan]$.
  Otherwise, $\AbstractPlan'$ will be added to the queue before line~\ref{alg:angelic:end_expand}.
  Therefore if $I_1(\PrimitivePlan)$ holds before line~\ref{alg:angelic:begin_expand}, it holds after line~\ref{alg:angelic:end_expand}.

  Initially, $Q = \{\textsc{Act}\}$.
  By construction, the operator $\TopLevelOperator$ contains every valid sequence of primitive operators.
  Therefore the proposition $\forall \PrimitivePlan: I_1(\PrimitivePlan)$ holds after line~\ref{alg:angelic:enqueue}.
  If $I_1(\PrimitivePlan)$ holds at line~\ref{alg:angelic:min}, it holds before line~\ref{alg:angelic:begin_expand}; therefore $\forall \PrimitivePlan: I_1(\PrimitivePlan)$ holds every time line~\ref{alg:angelic:check} is reached.

  If $\forall \PrimitivePlan: I_1(\PrimitivePlan)$ and $\hat{V}_U[\AbstractPlan^*] \prec \hat{V}_L[\AbstractPlan]\forall \AbstractPlan \in Q$, then $\forall \PrimitivePlan\exists \PrimitivePlan^* \in \AbstractPlan^*: V[\PrimitivePlan^*] \preceq V[\PrimitivePlan']$ and thus there exists an optimal primitive plan $\PrimitivePlan^* \in \AbstractPlan^*$.
  Then any time line~\ref{alg:angelic:success} is reached, $\AbstractPlan^*$ is an optimal primitive plan.
  In other words, if algorithm~\ref{alg:angelic} returns a plan, that plan is an optimal plan.

  If there is a feasible plan, it has finite cost, and therefore the cost $c^*$ of an optimal plan is finite.
  Let $N \in \mathbb{N}$ be the number of plans with lower bound less than $c^*$; because the abstraction is finite and every operator has a positive lower bound, $N$ is finite.
  Each plan is added to the queue at most once, and is therefore removed at most once.
  Then after $N$ iterations, the lower bound on any plan in $Q$ is greater than $c^*$, and algorithm~\ref{alg:angelic} will return a plan.



\end{proof}

\AcyclicOptimality
\begin{proof}
  Define the parent of a plan as 
  \begin{equation}
    \textsc{Parent}(\AbstractPlan') = \AbstractPlan \iff (\AbstractPlan, \AbstractPlan') \in \mathcal{R}).
  \end{equation}
  Define the ancestors of a plan recursively.
  \begin{align}
    \textsc{Ancestor}(\AbstractPlan, 0) &= \AbstractPlan \\
    \textsc{Ancestor}(\AbstractPlan, n) &= \nonumber\\
    &\hspace{-1cm}\textsc{Base}(\textsc{Parent}(\textsc{Ancestor}(\AbstractPlan, n-1))) \\
    \textsc{Ancestors}(\AbstractPlan) &= \{\textsc{Ancestor}(\AbstractPlan, n): n \in \Integers\}
  \end{align}
  Define the condition $I_2(Q, D, \PrimitivePlan)$ as true if one of
  \begin{itemize}
    \item $\exists \AbstractPlan \in Q : \PrimitivePlan \in \AbstractPlan$, or
    \item $\exists \AbstractPlan \in Q: \exists \AbstractPlan' \in \textsc{Def}(\AbstractPlan, D): \exists v \in \AbstractPlan: \PrimitivePlan \in  \circ \AbstractPlan'$, or
    \item $\exists \AbstractPlan \in Q: \hat{V}_U[\AbstractPlan] \prec V[\PrimitivePlan]$
  \end{itemize}
  is true.

  First, we show that if $I_2$ holds before removing a plan from the queue, it holds after the successors to that plan have been added to the queue.
  If $I_2(Q, D, \PrimitivePlan)$ holds before removing a plan $\AbstractPlan$, then either
  \begin{enumerate}
    \item $\exists \AbstractPlan' \ne \AbstractPlan \in Q: \PrimitivePlan \in \AbstractPlan'$
    \item $\exists \AbstractPlan' \ne \AbstractPlan \in Q: \AbstractPlan' \prec \PrimitivePlan $
    \item $\exists \AbstractPlan' \in Q: \exists \AbstractPlan'' \in \textsc{Anc}(\AbstractPlan', D): \exists \AbstractPlan''' \in \textsc{Def}(\AbstractPlan''): \PrimitivePlan \in \textsc{Base}(\AbstractPlan') \circ \AbstractPlan'''$, or
    \item $\PrimitivePlan \in \AbstractPlan$, or
    \item $\exists \AbstractPlan_a \in \textsc{Anc}(\AbstractPlan, D): \exists \AbstractPlan'' \in \textsc{Def}(\AbstractPlan'): \PrimitivePlan \in \AbstractPlan' = \textsc{Base}(\AbstractPlan) \circ \AbstractPlan''$
  \end{enumerate}
  In cases 1, 2, or 3, $\AbstractPlan' \in Q$ after expansion and the invariant holds.
  In case 4, because the abstraction is admissible, there is a plan $\AbstractPlan'$ such that $\PrimitivePlan \in \AbstractPlan'$ and $(\AbstractPlan, \AbstractPlan') \in \mathcal{R}$; this plan will be considered for expansion.
  In case 5, the plan $\AbstractPlan' = \textsc{Base}(\AbstractPlan) \circ \AbstractPlan''$ will be considered for expansion.
  If $\AbstractPlan'$ is acyclic, then either $\AbstractPlan' \in Q$ after expansion and the invariant holds, or $\AbstractPlan'$ is dominated by some plan already in $Q$ and the invariant holds.
  If $\AbstractPlan'$ is cyclic, its extension will be added to $\textsc{Def}(\textsc{Par}(\AbstractPlan))$.
  Since there is at least one refinement of $\AbstractPlan$ that will be added to the queue, there will be a plan on the queue with $\AbstractPlan$ as an ancestor, and thus the invariant will hold.

  Since $I_2$ holds initially, and holds after each expansion, it holds until the algorithm terminates.
  If $\forall \PrimitivePlan: I_2(Q, D, \PrimitivePlan)$ and $\hat{V}_U[\AbstractPlan^*] \prec \hat{V}_L[\AbstractPlan]\forall \AbstractPlan \in Q$, then $\forall \PrimitivePlan\exists \PrimitivePlan^* \in \AbstractPlan^*: V[\PrimitivePlan^*] \preceq V[\PrimitivePlan']$ and thus there exists an optimal primitive plan $\PrimitivePlan^* \in \AbstractPlan^*$.
  Then any time line~\ref{alg:angelic:success} is reached, $\AbstractPlan^*$ is an optimal primitive plan.
  In other words, if algorithm~\ref{alg:acyclic} returns a plan, that plan is an optimal plan.

  If there is a feasible plan, it has finite cost, and therefore the cost $c^*$ of an optimal plan is finite.
  Let $N \in \mathbb{N}$ be the number of acyclic plans with lower bound less than $c^*$; because the abstraction is finite and every acyclic plan has a cost strictly greater than its predecessor, $N$ is finite.
  Only acyclic plans are added to the queue; each plan is added at most once, and is therefore removed at most once.
  Then after $N$ iterations, the lower bound on any plan in $Q$ is greater than $c^*$, and algorithm~\ref{alg:angelic} will return a plan.

\end{proof}

\ApproximateOptimality
\begin{proof}
  Two invariants hold while the algorithm is running.
  First, invariant $I_2$ still holds; second, we know the minimal lower bound of any plan on the queue is less than $w$ times the lowest priority on the queue.
  \begin{align}
    \forall \PrimitivePlan: I_2(Q, D, \PrimitivePlan) \\
    \textsc{Key}(\AbstractPlan, w) \le w \hat{V}_L[\AbstractPlan] \forall \AbstractPlan \in Q
  \end{align}
  When the algorithm terminates,
  \begin{align}
    \hat{V}_U[\AbstractPlan^*](x_s, X_g) \prec \textsc{Key}(\AbstractPlan, w)\forall \AbstractPlan \in Q.
  \end{align}
  Consequently,
  \begin{align}
    \hat{V}_U[\AbstractPlan^*](x_s, X_g) \prec w \hat{V}_L[\AbstractPlan](x_s, X_g).
  \end{align}
  Since the algorithm must terminate by the same argument as for theorem~\ref{thm:acyclic:optimality}, algorithm~\ref{alg:approximate} will return a $w$-optimal solution in finite time.
\end{proof}

\end{document}